\documentclass[review]{elsarticle}

\usepackage{hyperref}
\usepackage{float}

\usepackage{algorithm} 
\usepackage{algorithmic} 
\usepackage{multirow} 
\usepackage{amsmath} 
\usepackage{xcolor}

\journal{ISPRS Journal of Photogrammetry and Remote Sensing }

\makeatletter
\newenvironment{breakablealgorithm}
  {
   \begin{center}
     \refstepcounter{algorithm}
     \hrule height.8pt depth0pt \kern2pt
     \renewcommand{\caption}[2][\relax]{
       {\raggedright\textbf{\ALG@name~\thealgorithm} ##2\par}%
       \ifx\relax##1\relax 
         \addcontentsline{loa}{algorithm}{\protect\numberline{\thealgorithm}##2}%
       \else 
         \addcontentsline{loa}{algorithm}{\protect\numberline{\thealgorithm}##1}%
       \fi
       \kern2pt\hrule\kern2pt
     }
  }{
     \kern2pt\hrule\relax
   \end{center}
  }
\makeatother

\begin{document}

\begin{frontmatter}

\title{SAR Image Classification Based on Spiking Neural Network through Spike-Time Dependent Plasticity and Gradient Descent}

\author[a,b,c,e]{Jiankun Chen}


\author[a,b,c,d]{Xiaolan Qiu\corref{mycorrespondingauthor}}
\cortext[mycorrespondingauthor]{Corresponding author}
\ead{xlqiu@mail.ie.ac.cn}

\author[a,c]{Chibiao Ding}
\author[c]{Yirong Wu}

\address[a]{National Key Laboratory of Science and Technology on Microwave Imaging, Institute of Electronics, Aerospace Information Research Institute, Beijing 10090, China}
\address[b]{The Key Laboratory of Technology in Geo-spatial Information Processing and Application System, Aerospace Information Research Institute, Chinese Academy of Sciences, Beijing 10090, China}
\address[c]{Aerospace Information Research Institute, Chinese Academy of Sciences, Beijing 100190, China}
\address[d]{Suzhou Research Institute, Aerospace Information Research Institute, Chinese Academy of Sciences, Suzhou 215123, China}
\address[e]{The University of Chinese Academy of Sciences, Beijing 10049, China}

\begin{abstract}
At present, the Synthetic Aperture Radar (SAR) image classification method based on convolution neural network (CNN) has faced some problems such as poor noise resistance and generalization ability. Spiking neural network (SNN) is one of the core components of brain-like intelligence and has good application prospects. This article constructs a complete SAR image classifier based on unsupervised and supervised learning of SNN by using spike sequences with complex spatio-temporal information. We firstly expound the spiking neuron model, the receptive field of SNN, and the construction of spike sequence. Then we put forward an unsupervised learning algorithm based on STDP and a supervised learning algorithm based on gradient descent. The average classification accuracy of single layer and bilayer unsupervised learning SNN in three categories images on MSTAR dataset is 80.8\% and 85.1\%, respectively. Furthermore, the convergent output spike sequences of unsupervised learning can be used as teaching signals. Based on the TensorFlow framework, a single layer supervised learning SNN is built from the bottom, and the classification accuracy reaches 90.05\%. By comparing noise resistance and model parameters between SNNs and CNNs, the effectiveness and outstanding advantages of SNN are verified. Code to reproduce our experiments is available at \url{https://github.com/Jiankun-chen/Supervised-SNN-with-GD}.  
\end{abstract}

\begin{keyword}
\texttt Spiking neural network (SNN)\sep SAR image classification\sep spike-time dependent plasticity (STDP) \sep gradient descent
\end{keyword}
\end{frontmatter}


\section{Introduction}

Artificial neural network (ANN) has made remarkable achievements and plays a crucial role in many fields such as image interpretation and voice processing. The first generation of ANN can be traced back to the McCulloch-Pitts (MP) \cite{1} model proposed in 1943, and its output is a Boolean logic variable. The second generation of ANN uses continuous activation functions \cite{2} to solve the problem of linear indivisibility. However, neither of them can simulate the membrane potential change and the spike emission process of biological neurons. The time information of a single spike is not directly used in the traditional ANN, of which the output is an analog value generally expressed in a given interval. In reality, the biological nervous system responds to various changes in the internal and external environment in the form of spike sequences \cite{3,4}, and more and more neuroscience researches show the importance of encoding and processing neural information based on precise spike timing \cite{5}. SNN comprises a more biologically interpretable spiking neuron model \cite{6}, which uses temporal coding \cite{7} for information transmission and processing. This coding method integrates multiple aspects of information, such as time, space, frequency, and phase. As the third generation of ANN, SNN has more powerful computing capabilities. Since SNN can simulate various neural signals and arbitrary continuous functions, it is an effective tool for complex spatio-temporal information processing \cite{8}.

The biological nervous system often uses different types of unsupervised learning, which means that the nervous system is plastic, and this plasticity is realized based on experience. The hippocampus plays a significant role in memory formation. Neural circuits can adjust or influence the topological structure and neural connection weights. Therefore, in a sense, unsupervised learning exists not only in a single neuron but also in large-scale neural networks. Recently, many unsupervised learning algorithms of SNNs have been proposed by researchers, which are used in data clustering and pattern recognition problems. Bohte et al. \cite{9} used SNN with spike time coding and Hebb learning rule to cluster high-dimensional data and effectively solve the classification problem of unsupervised remote sensing data. Diehl et al. \cite{10} improved the handwriting classification accuracy by utilizing the current-based synaptic model, STDP, lateral inhibition, and adaptive spiking threshold. The classification accuracy on the MINIST dataset is as high as 95$\%$. A series of comparative experiments have further verified the combination robustness of various mechanisms. Kheradpisheh et al. \cite{11} applied the STDP learning mechanism of an asynchronous feedforward SNN to invariant target recognition. The experimental results show that the model can accurately recognize different object instances in 3D-Object and ETH80 data sets.

The supervised learning algorithm of SNN is an emerging technology, and researchers have paid more and more attention to the supervised learning algorithm based on gradient descent nowadays. Experimental research shows that supervised learning exists in the biological nervous system, especially in the sensorimotor network and sensory system \cite{12}. When the sample situation changes, the synaptic weights can be modified to adapt to the new environment by supervised learning. Bohte et al. \cite{13} first put forward an error backpropagation suitable for multilayer feedforward SNNs, called SpikeProp algorithm, which uses a spike response model with analytical expression \cite{14}. Mckennoch et al. \cite{15} proposed RProp and QuickProp algorithms with faster convergence speed, further extended the SpikeProp algorithm to a nonlinear neuron model class and constructed the backpropagation algorithm of networks with Theta neuron \cite{16}. On this basis, Ghosh-Dastidar et al. \cite{17} deduced the gradient descent learning rule of synaptic weights by using chain rule. The authors put forward the Muli-SpikeProp algorithm and applied it to the standard XOR problem and more practical classification problem of Fisher Iris and electroencephalogram (EEG). Experimental results show that the Muli-SpikeProp algorithm has higher classification accuracy. Xu et al. \cite{18} put forward a new multi-spike supervised learning algorithm based on gradient descent. The algorithm has no limit on the spike number and realizes the spatio-temporal learning in multilayer feedforward SNN. The spike response model not only considers the spike sequence transmitted by presynaptic neurons but also sums up all the spikes sent by itself; that is, it has long-term memory characteristics. This model provides a proper theoretical basis for analyzing the dynamic characteristics and learning process of SNNs \cite{19,20}. Besides, Tino et al. \cite{21} extended the SpikeProp algorithm to SNNs with recursive structure.

The above research shows that the SNN’s penetration and application in image classification and other fields are still in a primary stage but promising. Synthetic Aperture Radar (SAR) image is a special type of image and has great difficulty in target classification and recognition. In recent years, deep learning provides a new tool for SAR image classification, which can learn the compelling features in SAR image data autonomously \cite{22,23}. However, the classification algorithms widely used in optical images such as Stacked Automatic Encoder (SAE) \cite{24}, Depth Trust Network (DBN) \cite{25}, and Convolutional Neural Network (CNN) \cite{26} faces some difficulties when applied to SAR images. The major problems are as follows:

(1) SAR images have inherent speckle noise. Speckle noise changes the gray value of SAR image pixels randomly, which reduces the validity and distinguishability of features and affects network performance on SAR image classification.

(2) The ambiguity and instability of targets’ characteristics in SAR images are serious. SAR images of the targets vary severely with the incident angle, azimuth angle, wavelength, and polarization. Moreover, there may be complicated interactions among different parts of the targets, resulting in severe phenomena of "the same object with different spectra" and "different objects with the same spectrum" \cite{27}. So, DNNs are difficult to learn robust and universal feature descriptors. 

In order to deal with these problems, this article tries to put the newly developed SNN forward into SAR image classification and studies its specific application methods. We take advantage of SNN in reasoning and judgment to improve the efficiency and accuracy of SAR image classification. The main contributions of this paper include:

(1) The application method of SNN in SAR image classification is explored for the first time. We rationally integrate the biological neuron model into the neural network. Experiments show that SNN can solve the problem of low classification accuracy caused by severe speckle noise and angle sensitivity.

(2) We further prove that the gradient backpropagation of SNN based on the use of optimizer is sufficient to update synapse weights. Then a single layer supervised learning SNN is innovatively built on TensorFlow platform. Experiments show that the single layer SNN is basically comparable with CNN in the task of SAR image classification.

The rest of the paper is arranged as follows: Chapter II introduces the essential elements of SNN; Chapter III elaborates the unsupervised learning algorithm based on STDP; Chapter IV further proposes the supervised learning algorithm based on gradient descent and via optimizer; Chapter V reports the constructed SNN for SAR image classification and provides the experimental results; Chapter VI analyzes the advantages of SNN in image noise resistance and model calculation by comparing with CNN. Finally, Chapter VII summarizes the whole paper.

\section{Related work}

Just as CNN is composed of convolution kernel, activation function, and other core parts, the elementary component of SNN are the spiking neuron, receptive field, and the construction of spike sequences. In order to facilitate subsequent analysis, some related works closely follow the research \cite{28} of Shikhar Gupta, University of Michigan. For the sake of clarity, the elementary components of our SNN for SAR image classification are described here.

\subsection{Spiking neuron}

Spiking neurons mimic small tiny nerve fibers in the brain, in which leaky integrated-and-fire (LIF) model \cite{29} has been widely used in the field of neural computing. In this model, the ion transfer of the biological nervous system is simulated by electron transfer, while the cell body is simulated by capacitance. The equivalent circuit of the LIF neuron model is shown in figure 1. A membrane capacitor $C_{m}$ and a membrane resistor $R_{m}$ are connected in parallel. When a presynaptic neuron emits a spike, a corresponding current $I(t)$ is generated on the synapse connected to the postsynaptic neuron. The current is divided into two parts, one part is used to charge the membrane capacitor $C_{m}$, and the other part flows away from the membrane resistor $R_{m}$.

\begin{figure}[ht!]
\begin{center}
		\includegraphics[width=0.6\columnwidth]{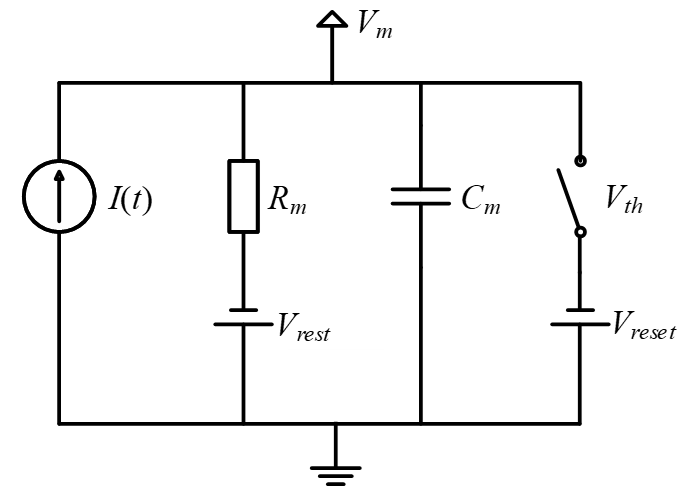}
	\caption{Equivalent circuit of LIF neuron model \cite{29}.}
\label{fig:figure_placement}
\end{center}
\end{figure}

The input current $I(t)$ is the weighted sum of synaptic currents generated by each presynaptic neuron's firing behavior. The current received by the postsynaptic neuron $i$ from the presynaptic neuron $j$ is expressed as formula (1).
\begin{equation}\label{equ:1}
	I(t) = \sum\limits_j {{\omega _{ij}}\sum\limits_f {\delta (t - t_j^f)} },
\end{equation}

Where $\omega_{ij}$ is the synaptic weight and  $t_j^f$ is the time when neuron $j$ emits the $f^{th}$ spike.

The first-order differential equation intuitively describes the relationship between the neuron’s membrane potential $V_{m}$ and its input current $I(t)$, as shown in formula (2) and formula (3).
\begin{equation}\label{equ:2}
	I(t) = \frac{{({V_m} - {V_{rest}})}}{{{R_m}}} + {C_m}\frac{{d{V_m}}}{{dt}},
\end{equation}
\begin{equation}\label{equ:3}
	{\tau _m}\frac{{d{V_m}}}{{dt}} =  - ({V_m} - {V_{rest}}) + {R_m}I(t),
\end{equation}

Where $\tau_{m}=C_{m}R_{m}$ is called membrane time constant. When the membrane potential $V_{m}$ is greater than the threshold potential $P_{th}$, neurons are excited immediately and generate an action potential (spike emission). Simultaneously, the membrane potential is reset to $V_{reset}$, which remains unchanged within the absolute refractory period $t_{ref}$. In other cases, the membrane potential will decay according to $\tau_{m}$ until the resting potential $V_{rest}$.

\subsection{Receptive field}

In biological vision, the receptive field innervated by one optic nerve fiber can be found from retina. The size of visual angle can generally express such visual receptive fields. They are composed of a central exciting field and a surrounding inhibitory field, called “On-center” cell structure, as shown in figure 2. Inspired by the above theory, the first layer neurons of SNN will be sensitive to different input image grids, having their specific receptive fields. The “On-center” cell structure is mutually concentric and circular. The exciting field and inhibitory field increase or decrease the spike transmission frequency, respectively.

\begin{figure}[ht!]
\begin{center}
		\includegraphics[width=0.7\columnwidth]{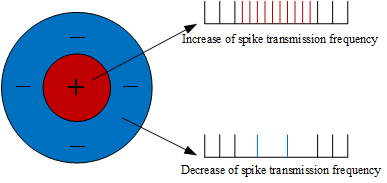}
	\caption{“On-center” cell structure.}
\label{fig:figure_placement}
\end{center}
\end{figure}

Taking into account the reduction of computational complexity, we try to implement SNN’s receptive field in a basic way. We propose a preliminary extraction method of SAR images based on a specific convolution kernel and Manhattan distance. Manhattan distance describes the distance between image grids. The convolution kernel weights are shown in the matrix (4) with a central symmetry structure. Compared with Euclidean distance, the calculation of Manhattan distance only involves addition and subtraction, which reduces the calculation cost and eliminates the error of taking approximations in square root calculations. The numerical setting of convolution kernel $\omega$ has the following characteristic: In the matrix, all the shortest paths formed by any two elements and on the premise of via the central element, traverse the same sum of elements value. This structure ensures the relative position of different image grids while filtering.
\begin{equation}\label{equ:4}
	\omega {\rm{  =  }}\left[ {\begin{array}{*{20}{c}}
{{\rm{ - 0}}{\rm{.5}}}&{{\rm{ - 0}}{\rm{.125}}}&{{\rm{0}}{\rm{.125}}}&{{\rm{ - 0}}{\rm{.125}}}&{{\rm{ - 0}}{\rm{.5}}}\\
{{\rm{ - 0}}{\rm{.125}}}&{{\rm{0}}{\rm{.125}}}&{{\rm{0}}{\rm{.5}}}&{{\rm{0}}{\rm{.125}}}&{{\rm{ - 0}}{\rm{.125}}}\\
{{\rm{0}}{\rm{.125}}}&{{\rm{0}}{\rm{.5}}}&{\rm{1}}&{{\rm{0}}{\rm{.5}}}&{{\rm{0}}{\rm{.125}}}\\
{{\rm{ - 0}}{\rm{.125}}}&{{\rm{0}}{\rm{.125}}}&{{\rm{0}}{\rm{.5}}}&{{\rm{0}}{\rm{.125}}}&{{\rm{ - 0}}{\rm{.125}}}\\
{{\rm{ - 0}}{\rm{.5}}}&{{\rm{ - 0}}{\rm{.125}}}&{{\rm{0}}{\rm{.125}}}&{{\rm{ - 0}}{\rm{.125}}}&{{\rm{ - 0}}{\rm{.5}}}
\end{array}} \right]\,
\end{equation}

Further, we obtain incentive images by convoluting the input images. Figure 3 shows the dynamic convolution process. Spike sequences guided by the incentive image pixels are loaded directly onto the first layer neurons. Therefore, each neuron corresponds to a fixed convolution region, that is, a specific receptive field of SNN.

\begin{figure}[ht!]
\begin{center}
		\includegraphics[width=1.0\columnwidth]{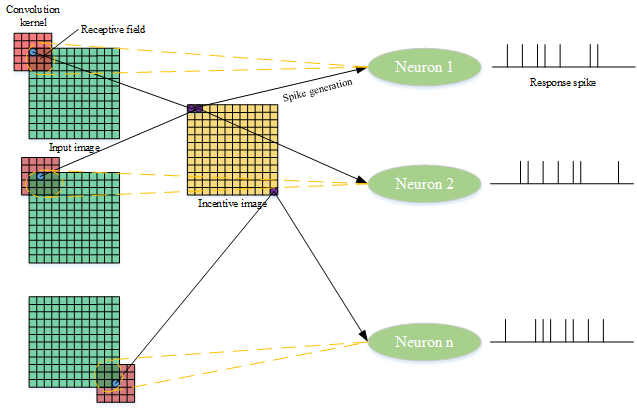}
	\caption{Dynamic convolution process in which the first layer neurons correspond to the receptive fields of input images. The yellow ellipses are receptive fields of SNN, and the blue circles are the centers of each receptive field.}
\label{fig:figure_placement}
\end{center}
\end{figure}

\subsection{Construction of spike sequences}

In SNN, the bio-interpretable information transmission is completed by precise timing spike sequences. The incentive image is an analog signal, which needs to be coded into spike sequences. Spike sequences are composed of discrete spike emission time. It is necessary to perform normalization on the incentive images to ensure that their pixel value ranges from 0 to 1. Timing information is determined by the spike frequency, which is proportional to the incentive images’ pixel value. We intend to design simple and practicable spike-based coding methods from the very beginning: random coding method, and deterministic coding method. These two methods can be applied to any images not only to SAR images. The specific implementation is as follows:

\begin{figure}[ht!]
\begin{center}
\par{
\begin{minipage}{1.0\textwidth}
\centering
\includegraphics[width=1.0\columnwidth]{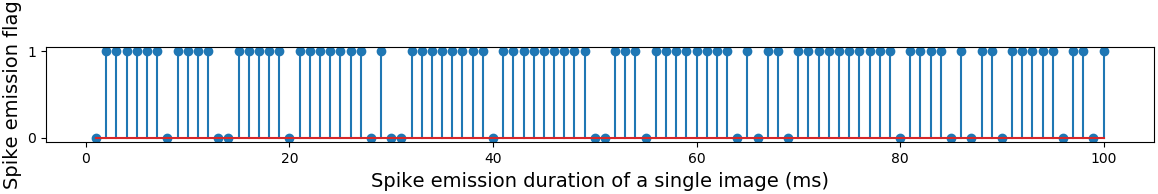}
\end{minipage}
}
\centerline{(a)}
\par{
\begin{minipage}{1.0\textwidth}
\centering
\includegraphics[width=1.0\columnwidth]{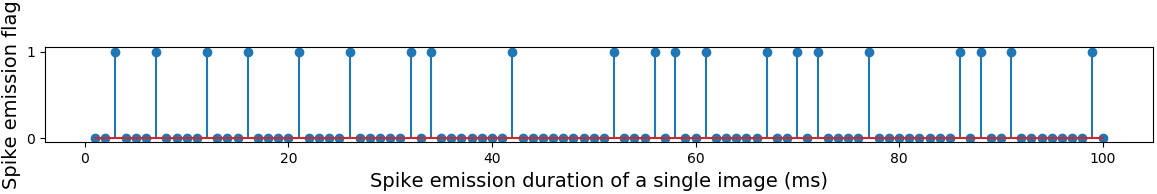}
\end{minipage}
}
\centerline{(b)}
\par{
\begin{minipage}{1.0\textwidth}
\centering
\includegraphics[width=1.0\columnwidth]{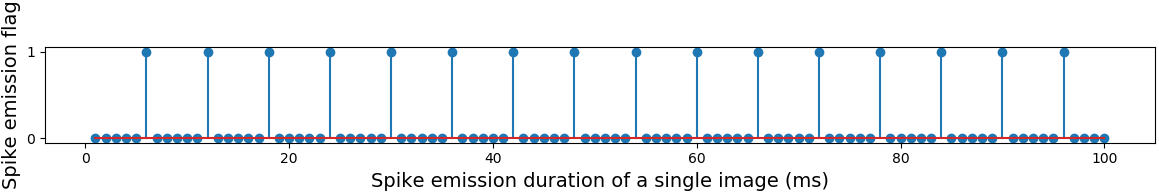}
\end{minipage}
}
\centerline{(c)}
\par{
\begin{minipage}{1.0\textwidth}
\centering
\includegraphics[width=1.0\columnwidth]{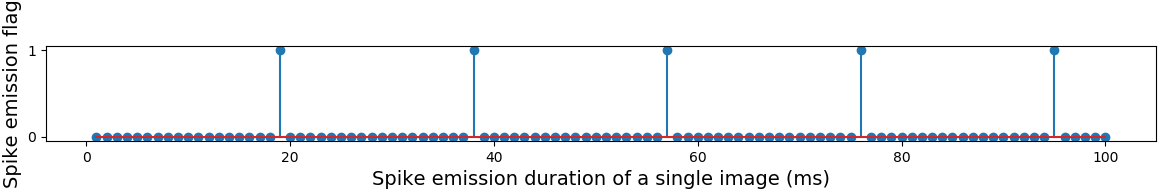}
\end{minipage}
}
\centerline{(d)}
\caption{Spike sequences guided by a incentive image of, (a) pixel value 0.788 based on the random coding method; (b) pixel value 0.212 based on the random coding method; (c) pixel value 0.788 based on the deterministic coding method; (d) pixel value 0.212 based on the deterministic coding method.}
\label{fig:figure_placement}
\end{center}
\end{figure}
\textbf{Random coding method:} For each pixel in the incentive image, we define a constant $T$ called spike emission duration of a single image (SEDSI) and divide the time axis 0 $\sim$ $T$ into $T$ parts according to time unit (TU). Encoder generates random numbers uniformly distributed within 0 $\sim$ 1. If pixel value is bigger than the random number, encoder will mark 1 at the memory and emit a spike; otherwise, encoder will mark 0 and keep silent. The above process can be summarized as calculating the probability of the spike number guided by each pixel. For example, when the pixel value is 0.788 and T=100 ms, spikes are emitted on about 78.8$\%$ of the 100 TU, as shown in the sub-picture (a) of figure 4. Similarly, the spike sequence guided by the complementary pixel value of 0.212 is shown in sub-picture (b).

\textbf{Deterministic coding method:} In each SEDSI, encoder generates uniform spike sequences according to a deterministic frequency. Since the average firing frequency of biological retinal neurons is between 10-200 Hz, the encoder emits spike 20 times in full frequency state and 1 time in the lowest frequency state. For the same pixel value of 0.788, a deterministic frequency $f_{det}$ =15.972 is linearly interpolated from the coordinates (0, 1) and (1, 20). We calculate the spike emission interval $I = \left\lfloor {T/{f_{\det }}} \right\rfloor  - 1$ on the time axis. Spike sequences guided by the pixel value of 0.801 and 0.199 based on the deterministic coding method are shown in the sub-picture (c) and (d) of figure 4, respectively.

\section{Unsupervised learning algorithm of SNN based on STDP}

In 1949, Donald Hebb \cite{32} discovered that altering the connection strength between neurons can complete nervous system learning called the Hebbian learning rule. Hebb learning rule is unsupervised. It specifies that the synaptic weight between two neurons should be increased or decreased in proportion to the product of their activation. The result of Hebb learning is that the network can extract the statistical characteristics of the training set and then divide the input data into several categories according to their similarity. To a certain extent, this is consistent with the process of humans observing and understanding the world. So, we firstly study the unsupervised learning algorithm for our SAR image classification SNN.

The neuroscience research found that the modifications of the connecting synapse are closely related to the precise time of neuron spike \cite{33}. Spike Timing Dependent Plasticity (STDP) is a temporally asymmetric form of Hebbian learning induced by tight temporal correlations between the spikes of presynaptic and postsynaptic neurons \cite{34}. The STDP learning mechanism is as follows: a presynaptic spike preceding a postsynaptic spike within a narrow time window leads to long-term potentiation (LTP); if the order is reversed, long-term depression (LTD) results.

The spike sequence sent by presynaptic neurons $i$ and postsynaptic neurons $j$ are expressed as formula (5):
\begin{equation}\label{equ:5}
{s_i}(t) = \sum\limits_f {\delta (t - t_i^f)}, \quad {s_j}(t) = \sum\limits_f {\delta (t - t_j^f)}
\end{equation}

With different parameters, formula (6) expresses various synaptic plasticity mechanisms \cite{36}. For the STDP learning rule, $C_0<0$, $C_1^{pre}>0$, and $C_1^{post}<0$.
\begin{equation}\label{equ:6}
\begin{array}{l}
\frac{{d{\omega _{ij}}(t)}}{{dt}} = {c_0} + {s_i}(t)[c_1^{pre} + \int_0^\infty  {{k^{pre,post}}(s){s_j}(t - s)ds} ]\\
{\rm{                  }} + {s_j}(t)[c_1^{post} + \int_0^\infty  {{k^{post,pre}}(s){s_i}(t - s)ds} ]
\end{array}
\end{equation}

In formula (6), every presynaptic or postsynaptic spike triggers the change of synaptic weight. $C_0$ represents a constant synaptic attenuation with independent nerve activity; that is, the synaptic weight gradually decreases in the absence of spikes. $C_1^{pre}$ and $C_1^{post}$ constitute the non-Hebbian term. The kernel functions $k^{pre,post}(s)$ and $k^{post,pre}(s)$ determine the shape of the learning window, where $s=t_j^{f}-t_i^{f}$ is the time difference between inter-spike interval (ISI) between presynaptic and postsynaptic spikes. The expression of STDP learning window is:
\begin{equation}\label{equ:7}
Window{(s)^{STDP}} = \left\{ {\begin{array}{*{20}{c}}
{{k^{post,pre}}(s) = {A_{+}}{e^{ - s/{\tau _ + }}},s \ge 0}\\
{{k^{pre,post}}(s) =  - {A_{-}}{e^{s/{\tau _ - }}},s < 0}
\end{array}} \right.
\end{equation}

Where, $A_{+}>0$ and $A_{-}>0$ are the maximum values of synaptic weight enhancement and inhibition respectively. $\tau_{+}$ and $\tau_{-}$ represent time constants respectively. Researchers got the STDP learning window, shown in figure 5. As a result of STDP, it is difficult to strengthen an enhanced synapse, and conversely, it is difficult to decline a weakened synapse. This can be applied to the SNNs’ learning model to ensure that the synaptic weights ${\omega _{ij}}$ are bounded \cite{37}. 

\begin{figure}[ht!]
\begin{center}
		\includegraphics[width=0.75\columnwidth]{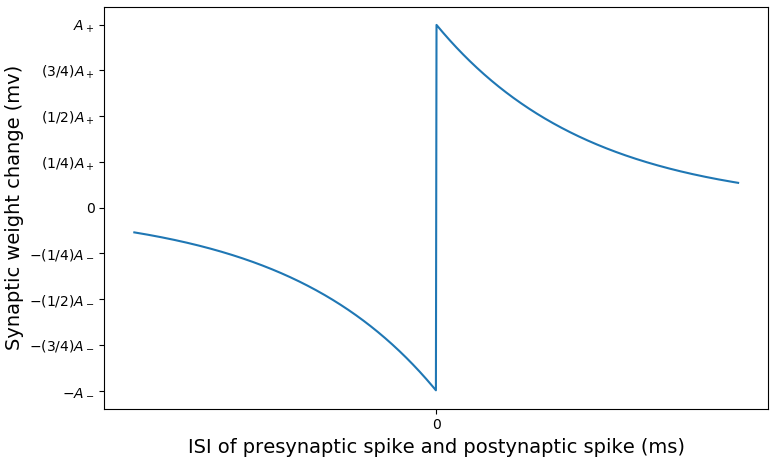}
	\caption{Alteration of synaptic weight versus ISI in the STDP learning window.}
\label{fig:figure_placement}
\end{center}
\end{figure}

Generally speaking, the synaptic weights in the initial stage are random. With the continuous feeding of input images, the membrane potentials change temporally from the rest state. The synaptic weights can be adjusted independently by the movement of neurons, and the change of neuronal connection in turn stimulates the posterior neuron, which echoes each other and achieves unsupervised learning.

In sensory physiology, lateral inhibition is a common theme that stimulated neurons inhibit the activity of nearby neurons. It takes place for preventing generated action potential’s spreading laterally, which further increases the sensory perception. The neuron that gets excited first inhibits (lowers down the membrane potential) of other neurons in the same layer. This property is called winner-takes-all (WTA). In vision, lateral inhibition enhances the contrast in brightness and facilitates edge detection. It is one of the SNN’s advantages to distinguish the vague edge between target and background, especially in SAR images.

\section{Supervised learning algorithm for SNN based on gradient descent}

Referring to the error back-propagation of the traditional ANN \cite{11}, gradient descent learning is to use the error between the expected neuron output and the actual output to obtain the gradient descent calculation result as the adjustment of synaptic weights. But for SNN, the difficulty of gradient descent is to overcome the discontinuity of membrane potential caused by spike firing \cite{13}. Supervised learning method is a frontier problem in the SNN research field. Here we propose a supervised learning method for SNN by directly using an optimizer. 

For the sake of brevity, let us consider a simple building block of SNNs, i.e., a spiking perceptron with $m$ presynaptic neurons and $n$ postsynaptic neurons. All the presynaptic neuron firing moments is defined as set $F$ defined in formula (8).
\begin{equation}\label{equ:8}
F = \sum\limits_{i = 1}^m {\left\{ {t_i^f{\rm{| }}p(t_i^f) \ge {P_{th}},{\rm{ }}t_i^f \ge 0} \right\}}
\end{equation}

Where $t_i^{f}$ is the moment when the presynaptic neuron $i$ sends out the $f^{th}$ spike. $P_{th}$ is the threshold potential. The potential $p(t)$ of the postsynaptic neuron is a function integrating the weighted presynaptic spikes as formula (9).
\begin{equation}\label{equ:9}
p(t) = \sum\limits_{i = 1}^m {\sum\limits_{{t_k} \in F} {{\omega _{ij}}g(t - {t_k})} }
\end{equation}

Where $g(·)$ is a given spike response function, and it is a continuously differentiable function deduced from the physical neuron model formula (3).

According to the gradient descent, the adjustment of synapse weight is calculated by the chain rule shown in formula (10). From now on, we stress the dependence of $p$ on $\omega$ and write $p(t,\omega )$ instead of $p(t)$.
\begin{equation}
\begin{array}{l}
\Delta {\omega _{ij}} =  - \eta \frac{{\partial E}}{{\partial {\omega _{ij}}}}
\\\quad \quad \; \, {\rm{ = }} - \eta \frac{{\partial E}}{{\partial p(t,\omega )}}\frac{{\partial p(t,\omega )}}{{\partial {\omega _{ij}}}}
\end{array}
\end{equation}

Where $\eta>0$ is the learning rate, and $p(t,\omega )$ is continuously differentiable with respect to $\omega$. Therefore, the error gradient back-propagation of the SNN can be carried out as long as the constructed error function $E$ is continuously differentiable to the membrane potential $p(t,\omega )$. 

The time $t$ and the synaptic weight $\omega$ are independent variables. For each small increment $\Delta \omega$ of $\omega$, we define a corresponding increment $\Delta t$ of $t$ by formula (11).
\begin{equation}\label{equ:11}
\Delta t = {\left( {\frac{{\partial p(t,\omega )}}{{\partial t}}} \right)^{ - 1}}\frac{{\partial p(t,\omega )}}{{\partial \omega }}\Delta \omega
\end{equation}

Although $p(t,\omega )$ is always a function of $t$, we can still be sure that the small increment of $p(t,\omega )$ brought by $t$ can be equivalent to that brought by $\omega$, to ensure that the gradient back-propagation process $\frac{{\partial p(t,\omega )}}{{\partial \omega }}$ in formula (10) is entirely free from the interference of variable $t$. The above conclusion is proved as formula (12). Since $p(t,\omega )$ is continuously differentiable with respect to $t$ and is linear with respect to $\omega$, we have:
\begin{equation}
\begin{array}{l}

\begin{array}{l}
p(t + \Delta t,\omega ) = p(t,\omega ) + \frac{{\partial p(t,\omega )}}{{\partial t}}\Delta t + o(\Delta t)\\\quad \quad \quad \quad \quad \,
{\rm{                   }} = p(t,\omega ) + \frac{{\partial p(t,\omega )}}{{\partial t}}{\left( {\frac{{\partial p(t,\omega )}}{{\partial t}}} \right)^{ - 1}}\frac{{\partial p(t,\omega )}}{{\partial \omega }}\Delta \omega  + o(\Delta t)\\\quad \quad \quad \quad \quad \, 
{\rm{                     }} = p(t,\omega ) + \frac{{\partial p(t,\omega )}}{{\partial \omega }}\Delta \omega  + o(\Delta t)\\\quad \quad \quad \quad \quad \, 
{\rm{                    }} = p(t,\omega  + \Delta \omega ) + o(\Delta t)\\\quad \quad \quad \quad \quad \, 
{\rm{                   }} \approx p(t,\omega  + \Delta \omega ) + o(\Delta \omega )
\end{array}

\end{array}
\end{equation}

This fact shows that it is sufficient to update the synaptic weights by using formula (9) in gradient back-propagation of SNN. Generally speaking, the supervised learning algorithm based on gradient descent is a mathematical analysis method. In the process of learning rule derivation, the state variables of SNN’s neuron model must have analytic expression, and linear models with fixed thresholds are mainly used, such as the LIF neuron model.

\section{Application of SNN in SAR images classification}

Based on the above components and learning algorithms for SNN, in this section, we built the SNN for SAR image classification and describes the detailed application method. We test the functional performance of the SNN based on the MSTAR dataset.

\subsection{MSTAR dataset}
The Moving and Stationary Target Acquisition and Recognition (MSTAR) dataset is one of the benchmark datasets \cite{41} for SAR image classification. The collected SAR image patches are 128 by 128 pixels with a resolution of one foot in range and azimuth. In order to explore the SNN performance of SAR image classification preliminarily, we opted for three categories of targets, including armored vehicle BMP-2, BTR-60, and tank T-72, shown in figure 6 on the basis of certain limitations of aircraft heading angle (list in table 1). The total sample number is 150, and there are 100 images in the training set and 50 images in the test set.
\begin{figure}[ht!]
\begin{center}
\par{
\begin{minipage}{0.25\textwidth}
\centering
\includegraphics[width=1.0\columnwidth]{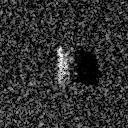}
\end{minipage}
\hspace{0.001\textwidth}
\begin{minipage}{0.25\textwidth}
\centering
\includegraphics[width=1.0\columnwidth]{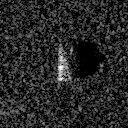}
\end{minipage}
\hspace{0.001\textwidth}
\begin{minipage}{0.25\textwidth}
\centering
\includegraphics[width=1.0\columnwidth]{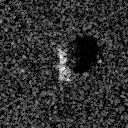}
\end{minipage}
}
\centerline{(a)}
\par{
\begin{minipage}{0.25\textwidth}
\centering
\includegraphics[width=1.0\columnwidth]{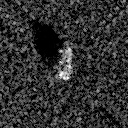}
\end{minipage}
\hspace{0.001\textwidth}
\begin{minipage}{0.25\textwidth}
\centering
\includegraphics[width=1.0\columnwidth]{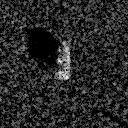}
\end{minipage}
\hspace{0.001\textwidth}
\begin{minipage}{0.25\textwidth}
\centering
\includegraphics[width=1.0\columnwidth]{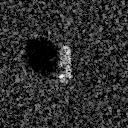}
\end{minipage}
}
\centerline{(b)}
\par{
\begin{minipage}{0.25\textwidth}
\centering
\includegraphics[width=1.0\columnwidth]{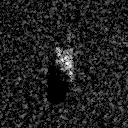}
\end{minipage}
\hspace{0.001\textwidth}
\begin{minipage}{0.25\textwidth}
\centering
\includegraphics[width=1.0\columnwidth]{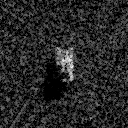}
\end{minipage}
\hspace{0.001\textwidth}
\begin{minipage}{0.25\textwidth}
\centering
\includegraphics[width=1.0\columnwidth]{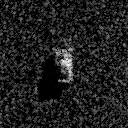}
\end{minipage}
}
\centerline{(c)}
\caption{Three categories of images of MSTAR dataset. (a) BMP-2, (b) BTR-60, and (c) T-72.}
\label{fig:figure_placement}
\end{center}
\end{figure} 

\begin{table}[H]
	\centering
		\begin{tabular}{|l|c|c|c|}\hline
			Category&BMP-2&BTR-60&T-72\\\hline
			 Minimum aircraft heading angle&13.191°&7.483°&-49.288°\\\hline
			 Maximum aircraft heading angle&73.191°&75.483°&10.712°\\\hline
			 Training sample number&100&100&100\\\hline
			 Test sample number&50&50&50\\\hline
		\end{tabular}
	\caption{Partial MSTAR dataset.}
\label{tab:Partial MSTAR dataset}
\end{table}

\subsection{SAR image classification based on unsupervised learning of SNN}
\subsubsection{Unsupervised learning algorithm of the single layer SNN}
 In this article, a single layer SNN with the function of three categories of image classification is firstly proposed, as shown in figure 7. It is a very simple feedforward SNN. There are 128 × 128 neurons in the input layer and three neurons in the output layer. Firstly, SNN converts the input images into incentive images through a 5 × 5 receptive field. Then the spike generator transmits spiking sequences and directly maps them to the input layer pixel by pixel. The neuronal dynamics complies with the LIF model characteristics. Synapse weights are passively updated according to STDP to complete continuous learning of SNN. Due to the lateral inhibition mechanism, a unique neuron in the output layer generates an action potential (in the case of SNN convergence) in each SEDSI. The serial number of this response neuron represents the image category.
\begin{figure}[H]
\begin{center}
		\includegraphics[width=0.94\columnwidth]{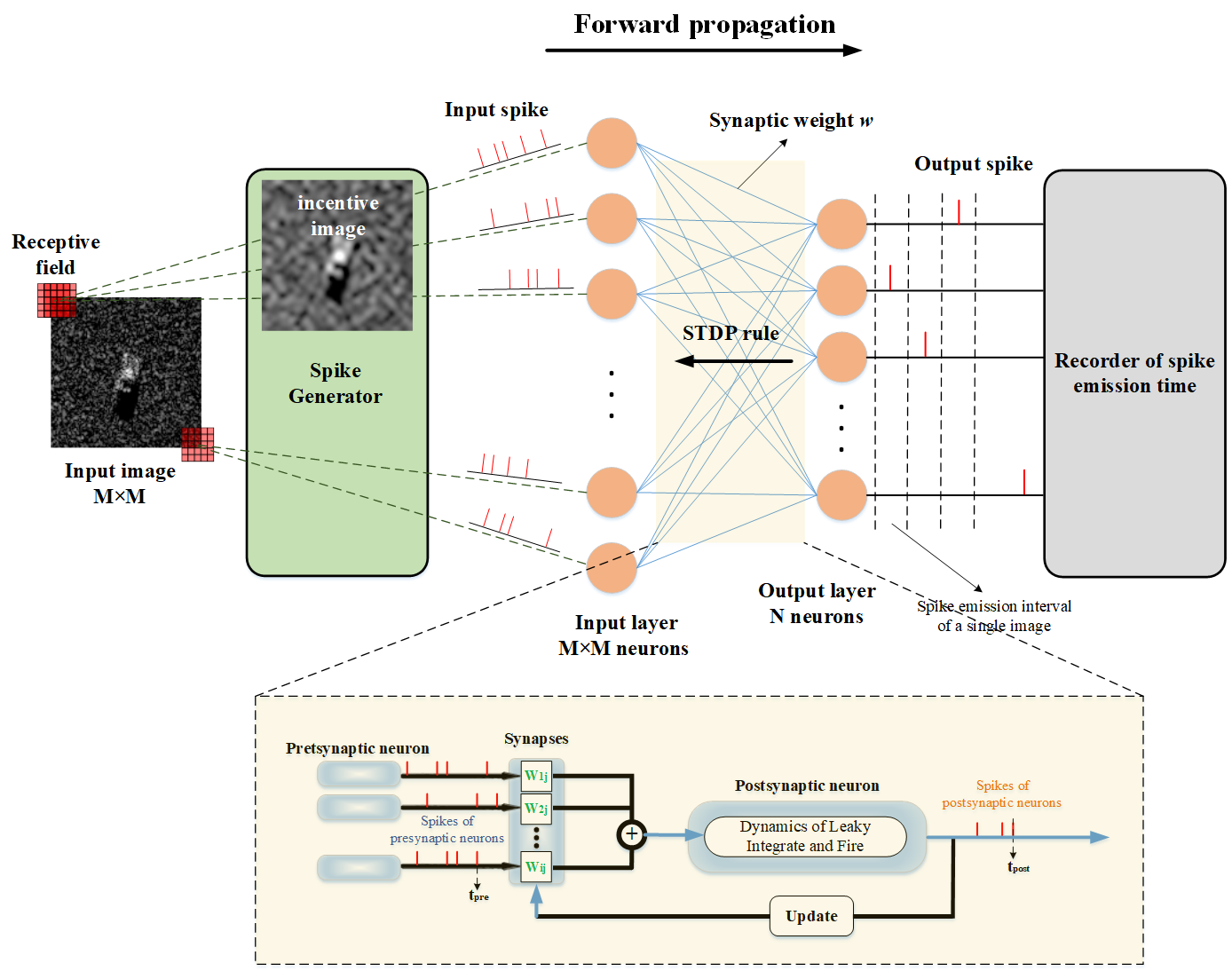}
	\caption{Architecture of unsupervised learning single layer feedforward SNN.}
\label{fig:figure_placement}
\end{center}
\end{figure}

The unsupervised learning algorithm is shown in algorithm 1. Since the neurons in the input layer provide spike-based encoding of sensory inputs, the postsynaptic neuron change membrane potential and cause an action potential. SNN updates synaptic weights based on STDP, which makes the synaptic weights converge to a stable state after more and more images are input. This process is the image feature extraction by SNN.
\\
\renewcommand{\algorithmicrequire}{\textbf{Input:}} 
\renewcommand{\algorithmicensure}{\textbf{Output:}}
\begin{breakablealgorithm} 
\caption{Unsupervised learning of the single layer feedforward SNN} 
\label{alg1} 
\begin{algorithmic}[0] 
\REQUIRE Unpreprocessed image patches 
\ENSURE Prediction of input image category and classification accuracy 
\STATE \textbf{1. Preparation of SNN’s elementary components:}
\STATE 1.1 create the time unit (TU) axis $0, 1, 2, ... T$ according to the SEDSI $T$
\STATE 1.2 establishe LIF neurons in the input and output layer
\STATE 1.3 set up a convolution kernel $\omega$ as the receptive field
\STATE 1.4 build a spike generator and a recorder of spike emission time
\STATE 1.5 initialize synaptic weights 
\STATE 1.6 fix the parameters for forward or backward spike search in STDP: $t_{fore}$ and $t_{back}$
\STATE \textbf{2. Unsupervised training of SNN:}
\FOR{each epoch} 
\FOR{each input image}
\STATE 2.1 generate an incentive image and spike sequence
\STATE 2.2 initialize membrane potential
\STATE 2.3 forward propagation of SNN:
\FOR{each $TU$}
\WHILE{the current time is not in the refractory period} 
\STATE 2.3.1 carry out the LIF neuronal dynamic
\STATE 2.3.2 generate action potentials:
\IF{the threshold potential is reached} 
\STATE (i) emit a spike, return the membrane potential to the resting potential, and  execute lateral inhibition
\STATE (ii) record the current time as $t_{spike}$
\STATE (iii) generate a refractory period
\STATE (iv) record the neuron that emits spike as “winner” (continuously update with the traversal of $TU$)
\ELSE
\STATE enter the next $TU$ and return to 2.3.1-2.3.2
\ENDIF
\ENDWHILE 
\ENDFOR
\STATE 2.4 update synaptic weights based on STDP:
\FOR{each $t_{spike}$}
\FOR{each neuron in the output layer}
\WHILE{it has emitted a spike at $t_{spike}$}
\FOR{each presynaptic neuron}
\IF{the spike emission is in the period ($t_{spike}$-$t_{fore}$, $t_{spike}$)}
\STATE strengthen the connected synaptic weight
\ELSIF{the spike emission is in the period ($t_{spike}$, $t_{spike}$+$t_{back}$) } 
\STATE weaken the connected synaptic weight 
\ELSE 
\STATE do not update the connected synaptic weight 
\ENDIF 
\ENDFOR
\ENDWHILE
\ENDFOR
\ENDFOR
\STATE 2.5 micro modify synaptic weights
\FOR{each neuron in the input layer}
\IF{it has not emitted any spike in the SEDSI $T$}
\STATE reduce the connected synaptic weight between it and the “winner” by 0.2mv
\ENDIF
\IF{the connected synaptic weight between it and the “winner” is less than the lower limit $\omega_{min}$}
\STATE reset the synaptic weight to $\omega_{min}$
\ENDIF
\ENDFOR
\ENDFOR
\ENDFOR
\STATE \textbf{3. Test of SNN:}
\STATE 3.1 input the test image into SNN only for forward propagation based on the pretrained synaptic weights
\STATE 3.2 output the prediction of image category, i.e. the serial number of the “winner”
\STATE 3.3 calculate the classification accuracy
\end{algorithmic} 
\end{breakablealgorithm}

\subsubsection{Unsupervised learning algorithm of the bilayer SNN}

Based on section 5.2.1, a bilayer feedforward SNN with a hidden layer is further built. As shown in figure 8, the update of synapse weight is completed from back to front according to the two STDPs. However, the implementation of STDP is different between the bilayer and single-layer SNNs. The reason is: in the single-layer SNN, the input layer neurons provide spike-based encoding of sensory inputs and do not implement the neuronal dynamics. We can directly obtain the membrane potential matrix of the input layer neurons in the entire SEDSI. The search of pre and postsynaptic correlation spike is performed for each TU (1ms), and the updated weights are used immediately for the next TU training. However, the weight updating of the output layer needs to input the membrane potential matrix of the hidden layer neurons, which cannot be calculated instantaneously but takes time to accumulate the current TU until the complete SEDSI. Therefore, for the bilayer feedforward SNN, we propose an approximate real-time synaptic weight updating method to seek the optimal result, illustrated in algorithm 2. We set SEDSI as 70ms. In each SEDSI, synapse weights are updated 10 times (every 7ms).
\\
\begin{figure}[ht!]
\begin{center}
		\includegraphics[width=1.0\columnwidth]{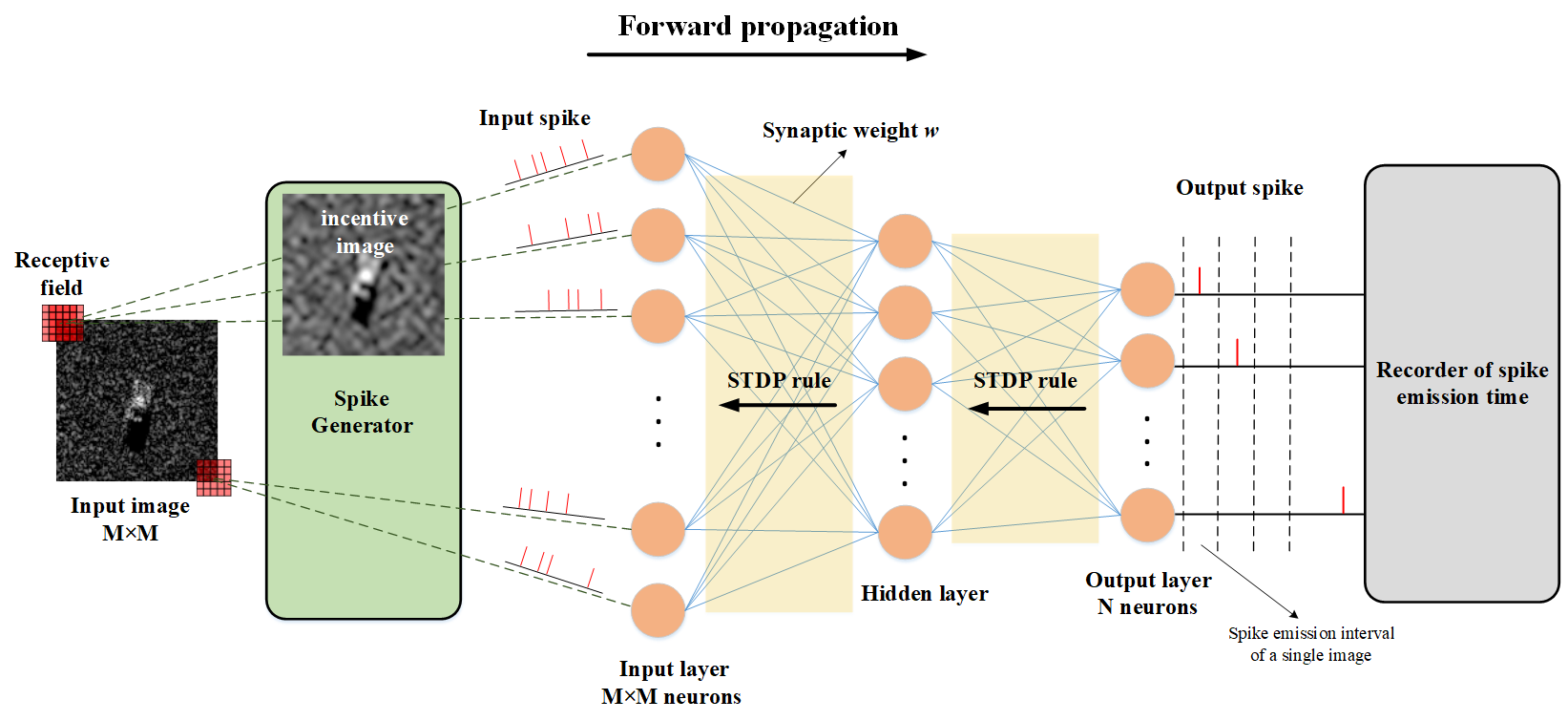}
	\caption{Architecture of unsupervised learning bilayer feedforward SNN.}
\label{fig:figure_placement}
\end{center}
\end{figure}

\renewcommand{\algorithmicrequire}{\textbf{Input:}} 
\renewcommand{\algorithmicensure}{\textbf{Output:}}
\begin{breakablealgorithm} 
\caption{An approximate real-time synaptic weight updating method for bilayer SNN} 
\label{alg1} 
\begin{algorithmic}[0] 
\REQUIRE Unpreprocessed image patches 
\ENSURE Prediction of input image category and classification accuracy 
\STATE \textbf{1. Create the variables and memory space:}
\STATE 1.1 create matrix variables, including membrane potential of the input layer $act\_pot\_0$, membrane potential of the hidden layer $act\_pot\_1$, membrane potential of the output layer $act\_pot\_ 2$, synapse  weights between the input layer and the hidden layer $syn\_01$, synapse weights between the hidden layer and the output $syn\_ 12$
\STATE 1.2 divide SEDSI into 10 subsegments uniformly and record them as $\Delta t_{1}$, $\Delta t_{2}$ …, $\Delta t_{10}$
\STATE \textbf{2. Update synaptic weights :}
\STATE 2.1 enter the first subsegment $\Delta t_{1}$, record $act\_pot\_2(\Delta t_{1})$ and $act\_pot\_1(\Delta t_{1})$
\STATE 2.2 apply STDP between the output layer and the hidden layer:
\IF{the maximum value in $act\_pot\_2(\Delta t_{1})$ exceeds $P_{th}$}
\STATE output the moment of the maximum value as $TU_{max}$, and search the correlation spike in the period ($TU_{max}$ -3.5, $TU_{max}$ +3.5) in $act\_pot\_1(\Delta t_{1})$, and the update $syn\_ 12$ by STDP
\ELSE 
\STATE do not update $synapse\_ 12$, and go to 2.3 
\ENDIF 
\STATE 2.3 apply STDP between the hidden layer and the input layer:
\IF{the maximum value in $act\_pot\_2(\Delta t_{1})$ exceeds $P_{th}$}
\STATE output the moment of the maximum value as $TU_{max}$, and search the correlation spike in the period ($TU_{max}$ -3.5, $TU_{max}$ +3.5) in $act\_pot\_0(\Delta t_{1})$, and the update $syn\_ 01$ by STDP
\ELSE 
\STATE do not update $syn\_ 01$, and go to 2.4
\ENDIF
\STATE 2.4	enter the next subsegments $\Delta t_{2}$ and repeat steps 2.2 $\sim$ 2.3 until $\Delta t_{10}$
\end{algorithmic} 
\end{breakablealgorithm}

In order to verify the necessity of the approximate real-time synaptic weight updating method, we have carried out an experiment using the non-real-time synaptic weight updating method (synaptic weights are updated only once within a SEDSI). With iteration increment, neurons are silent in the entire SEDSI of more and more input images. Finally, the SNN only updates synaptic weights for three specific images that evolved into the feature maps. In this case, SNN uses an individual image to represent the feature of one-category images, resulting in classification errors. To solve this problem, we propose a bilayer SNN based on the approximate real-time synaptic weight updating method and outperform the single-layer SNN in terms of convergence speed and classification accuracy.

\subsubsection{Classification results and feature visualization of synaptic weights}
\begin{figure}[ht!]
\begin{center}
\par{
\begin{minipage}{0.9\textwidth}
\centering
\includegraphics[width=1.0\columnwidth]{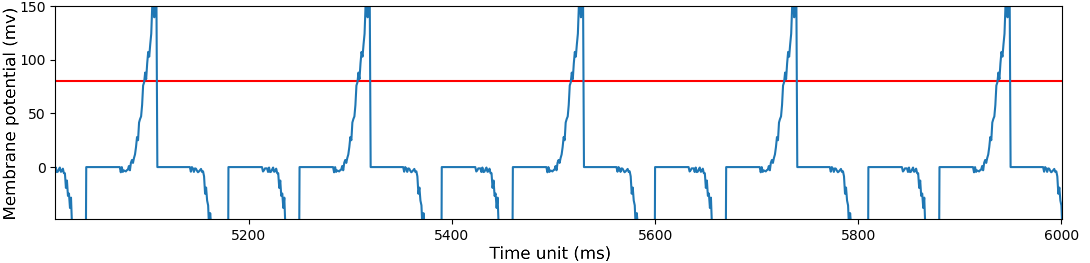}
\end{minipage}
}
\centerline{(a)}
\par{
\begin{minipage}{0.9\textwidth}
\centering
\includegraphics[width=1.0\columnwidth]{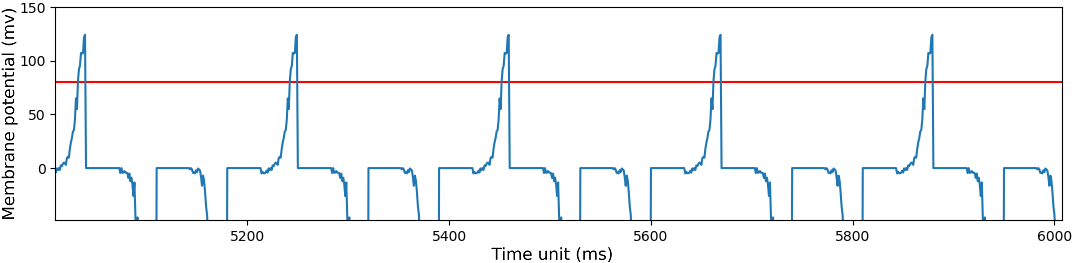}
\end{minipage}
}
\centerline{(b)}
\par{
\begin{minipage}{0.9\textwidth}
\centering
\includegraphics[width=1.0\columnwidth]{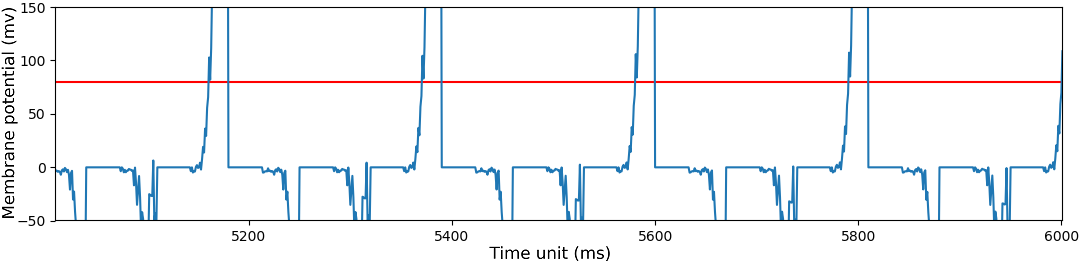}
\end{minipage}
}
\centerline{(c)}
\caption{Response of output layer neurons when three categories of images are input in turn, (a) BMP-2, (b) BTR-60, and (c) T-72. The blue curves are membrane potential, and the red lines are threshold potential.}
\label{fig:figure_placement}
\end{center}
\end{figure}
The reorder of spike emission time gives out the winner neuron in the output layer. Hence, we can determine the categories to which input images belong. We consider that SNN has reached a convergence state when the classification accuracy stagnates or fluctuates up and down. Figure 9 shows the membrane potential of the output layer neurons in the converged SNN. When we input the three categories of images, in turn, the three neurons in the output layer respond alternately.

The learning process of SNN can also be expressed by the curve of classification accuracy versus training epoch, as shown in figure 10. The numerical values are given in table 2.
\begin{figure}[ht!]
\begin{center}
\begin{minipage}{1.0\textwidth}
\centering
\includegraphics[width=1.0\columnwidth]{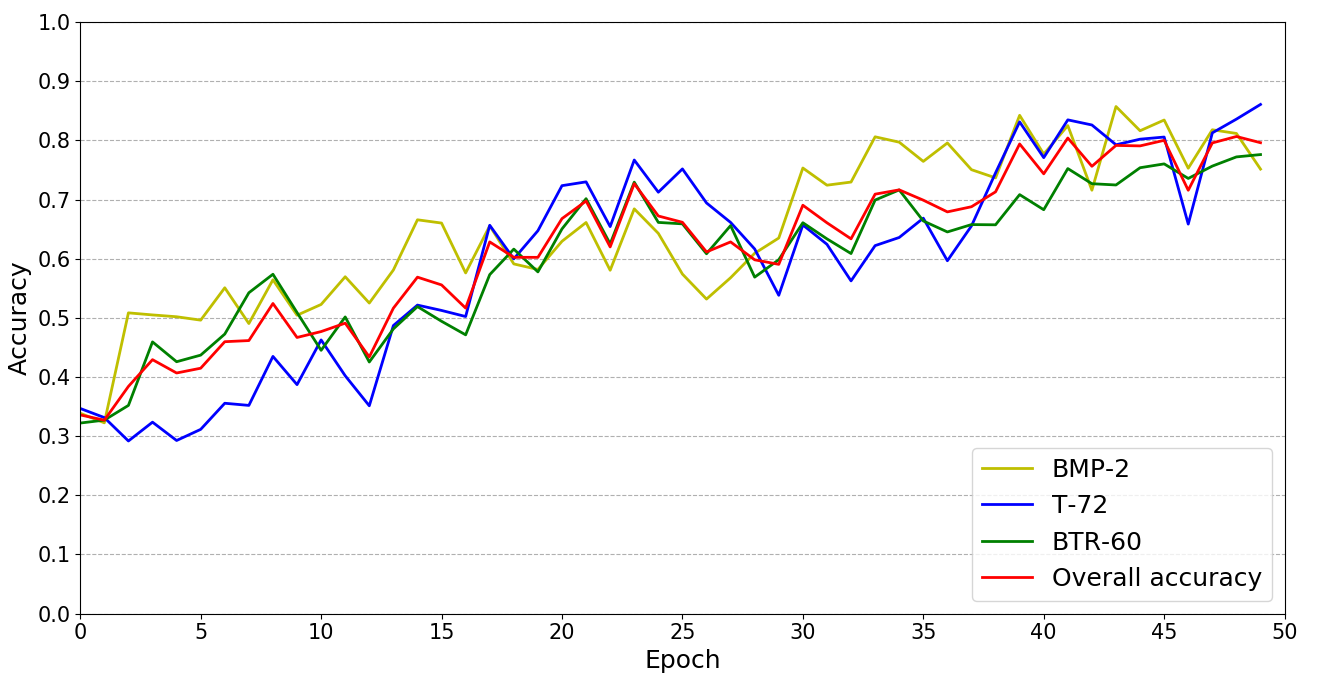}
\centerline{(a)}
\end{minipage}
\hspace{0.001\textwidth}
\begin{minipage}{1.0\textwidth}
\centering
\includegraphics[width=1.0\columnwidth]{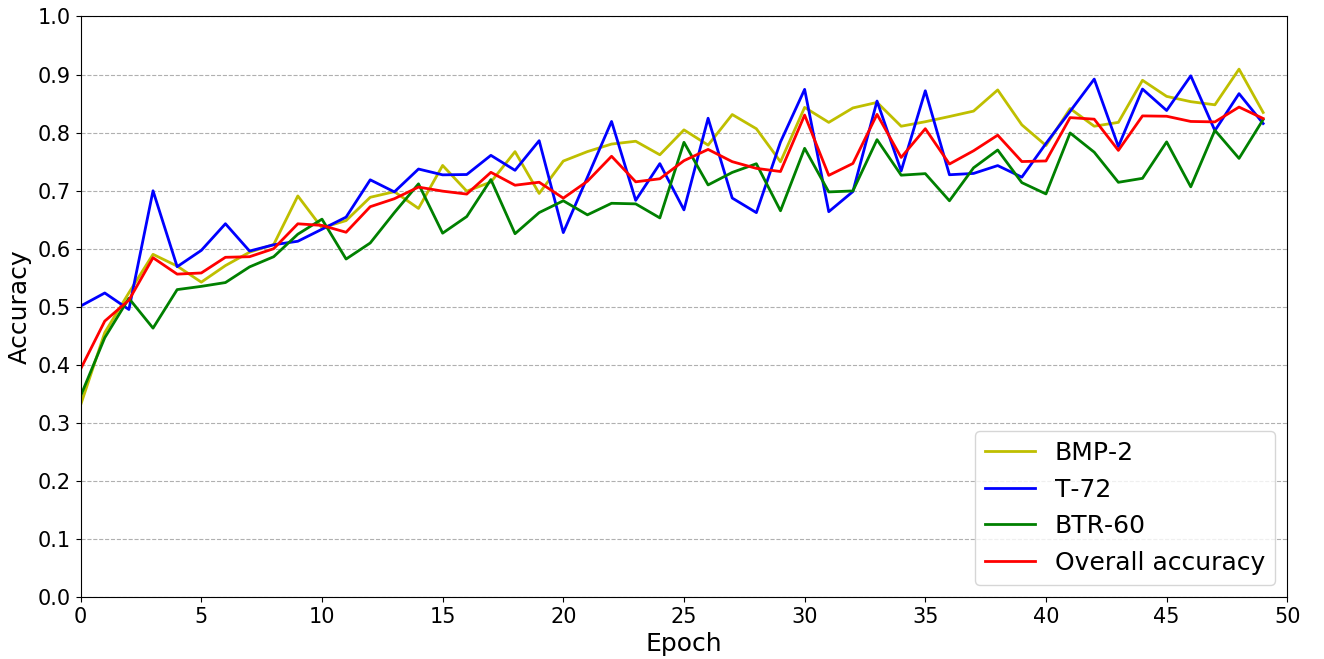}
\centerline{(b)}
\end{minipage}
\caption{Classification accuracy of three categories SAR images versus training epoch based on the unsupervised learning SNN, (a) single-layer SNN; (b) bilayer SNN.}
\label{fig:figure_placement}
\end{center}
\end{figure}

\begin{table}[H]
	\centering
		\begin{tabular}{|l|c|c|c|c|}\hline
			Category&BMP-2&BTR-60&T-72&Overall accuracy\\\hline
			 Classification accuracy of&0.815&0.773&0.836&0.808\\
			 single-layer SNN& & & & \\\hline
			 Classification accuracy of&0.881&0.803&0.867&0.851\\
			 bilayer SNN& & & & \\\hline
		\end{tabular}
	\caption{Classification accuracy of three categories SAR images based on the unsupervised learning SNN.}
\label{tab:Classification accuracy of three categories SAR images based on the unsupervised learning SNN}
\end{table}

By rearranging the synaptic weights according to the receptive field, we visualize feature maps of the unsupervised learning SNN, as shown in figure 11 and figure 12.
\begin{figure}[ht!]
\begin{center}
\par{
\begin{minipage}{0.31\textwidth}
\centering
\includegraphics[width=1.0\columnwidth]{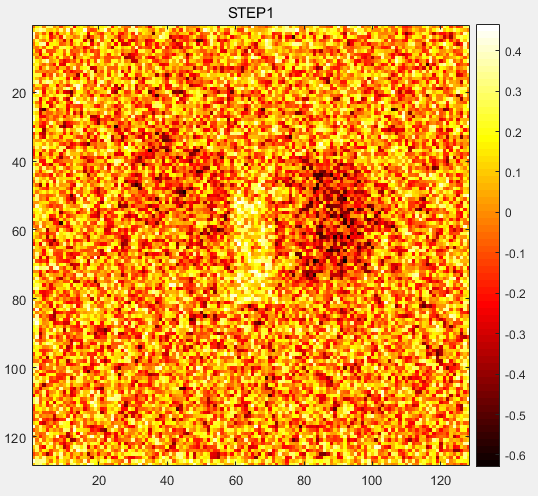}
\end{minipage}
\hspace{0.001\textwidth}
\begin{minipage}{0.31\textwidth}
\centering
\includegraphics[width=1.0\columnwidth]{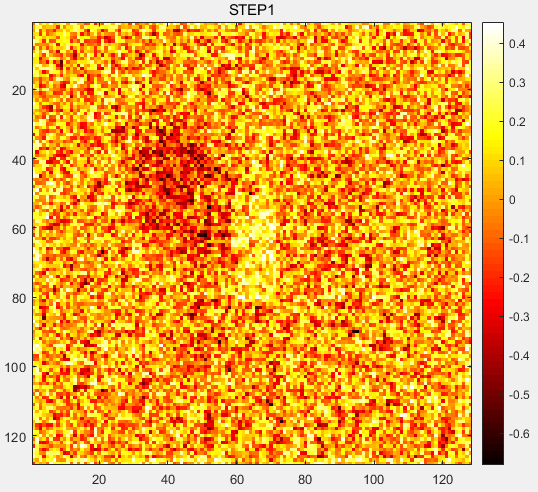}
\end{minipage}
\hspace{0.001\textwidth}
\begin{minipage}{0.31\textwidth}
\centering
\includegraphics[width=1.0\columnwidth]{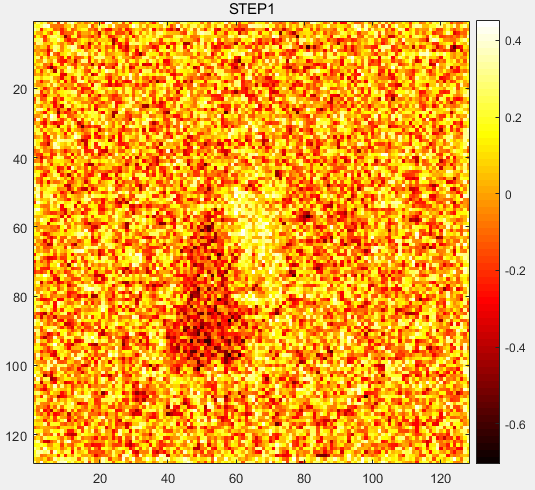}
\end{minipage}
}
\centerline{(a)}
\par{
\begin{minipage}{0.31\textwidth}
\centering
\includegraphics[width=1.0\columnwidth]{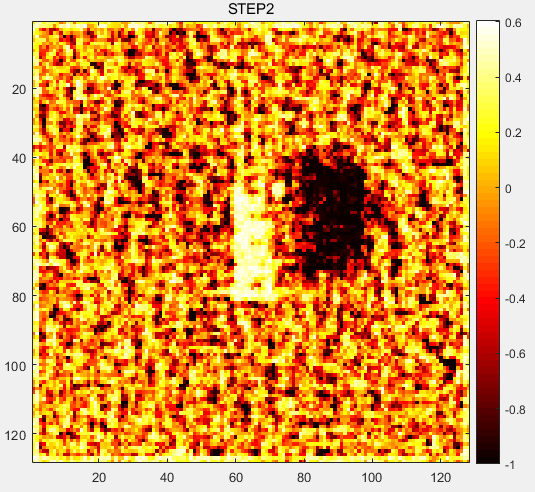}
\end{minipage}
\hspace{0.001\textwidth}
\begin{minipage}{0.31\textwidth}
\centering
\includegraphics[width=1.0\columnwidth]{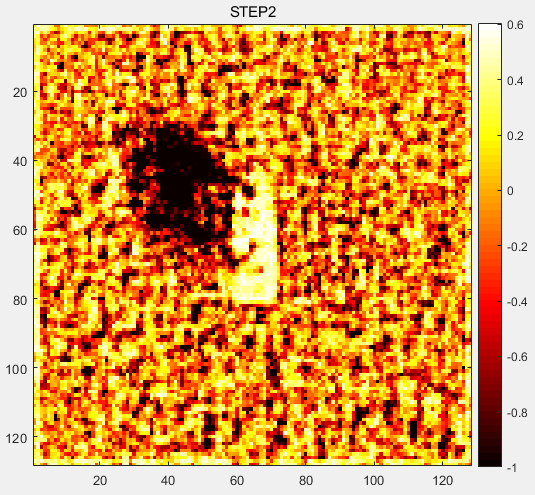}
\end{minipage}
\hspace{0.001\textwidth}
\begin{minipage}{0.31\textwidth}
\centering
\includegraphics[width=1.0\columnwidth]{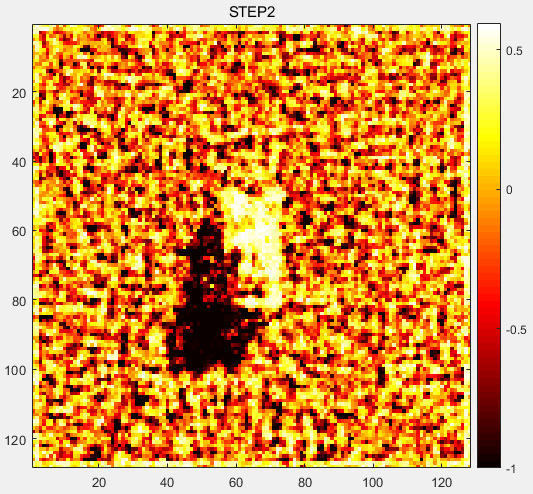}
\end{minipage}
}
\centerline{(b)}
\par{
\begin{minipage}{0.31\textwidth}
\centering
\includegraphics[width=1.0\columnwidth]{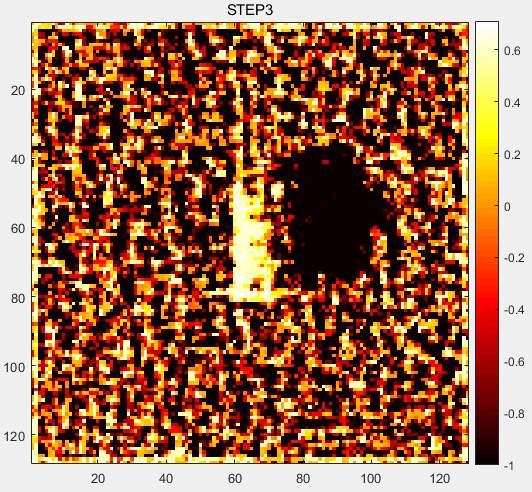}
\end{minipage}
\hspace{0.001\textwidth}
\begin{minipage}{0.31\textwidth}
\centering
\includegraphics[width=1.0\columnwidth]{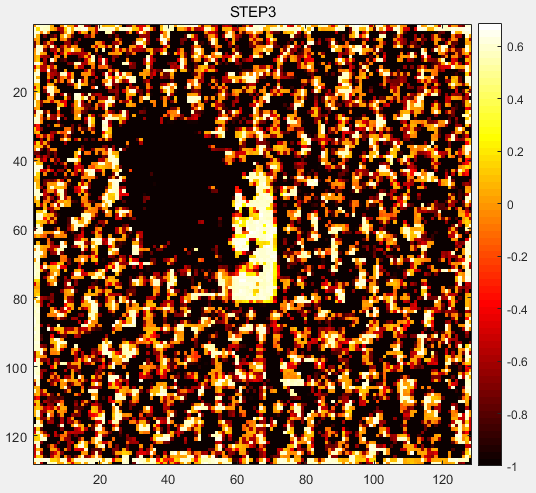}
\end{minipage}
\hspace{0.001\textwidth}
\begin{minipage}{0.31\textwidth}
\centering
\includegraphics[width=1.0\columnwidth]{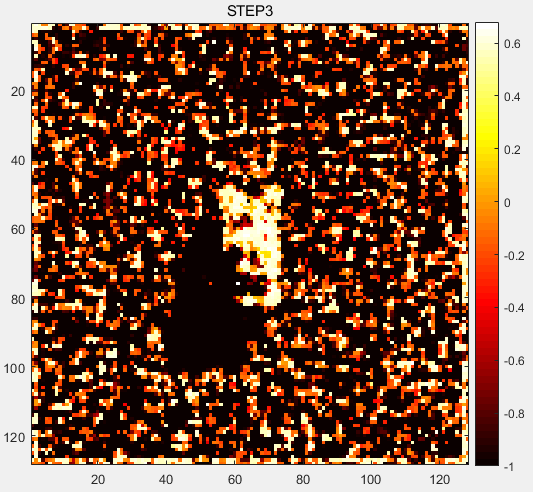}
\end{minipage}
}
\centerline{(c)}
\caption{Visualization of synaptic weights in the unsupervised learning single-layer SNN where each column from left to right is the feature map of BMP-2, BTR-60, T-72 via (a) 2, (b) 6, and (c) 20 training epochs, respectively.}
\label{fig:figure_placement}
\end{center}
\end{figure} 
\begin{figure}[H]
\begin{center}
\par{
\begin{minipage}{0.31\textwidth}
\centering
\includegraphics[width=1.0\columnwidth]{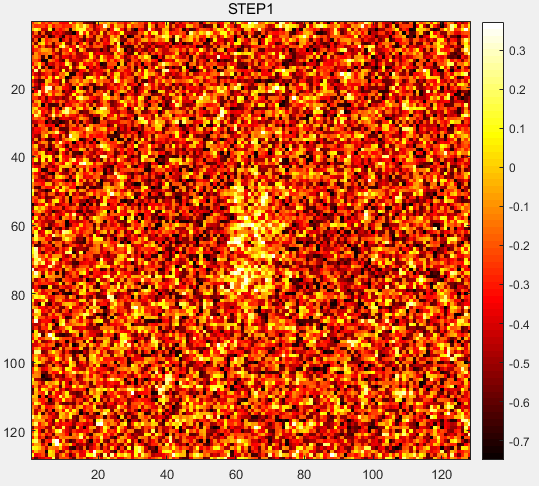}
\end{minipage}
\hspace{0.001\textwidth}
\begin{minipage}{0.31\textwidth}
\centering
\includegraphics[width=1.0\columnwidth]{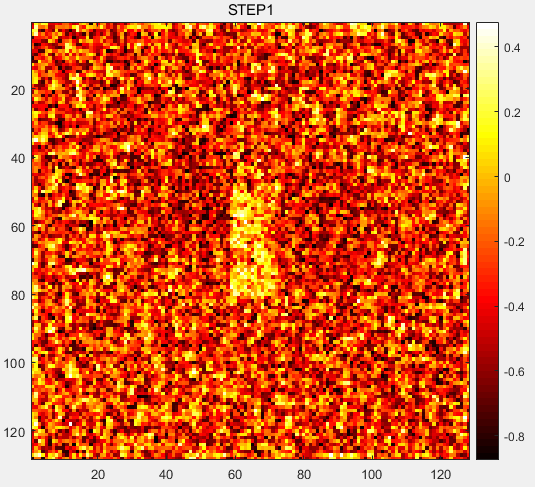}
\end{minipage}
\hspace{0.001\textwidth}
\begin{minipage}{0.31\textwidth}
\centering
\includegraphics[width=1.0\columnwidth]{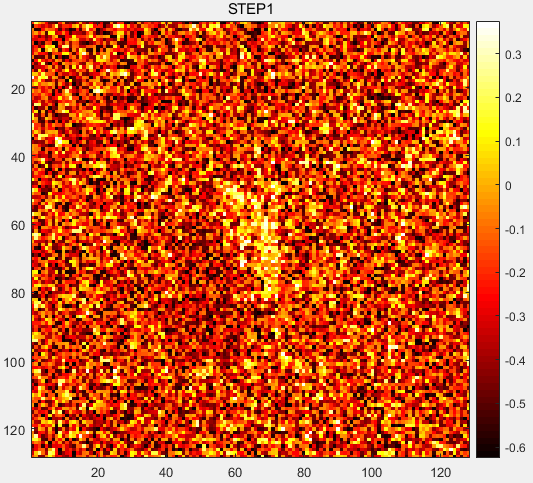}
\end{minipage}
}
\centerline{(a)}
\par{
\begin{minipage}{0.31\textwidth}
\centering
\includegraphics[width=1.0\columnwidth]{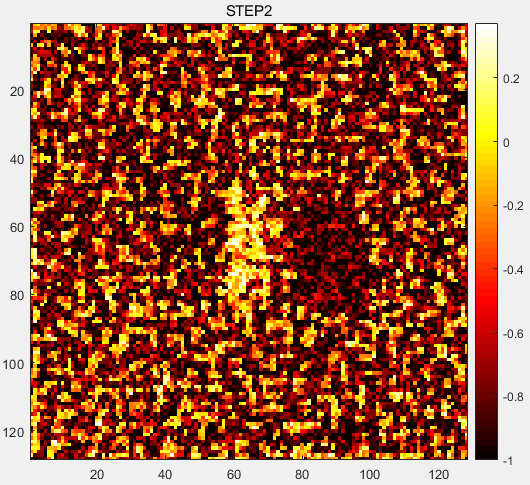}
\end{minipage}
\hspace{0.001\textwidth}
\begin{minipage}{0.31\textwidth}
\centering
\includegraphics[width=1.0\columnwidth]{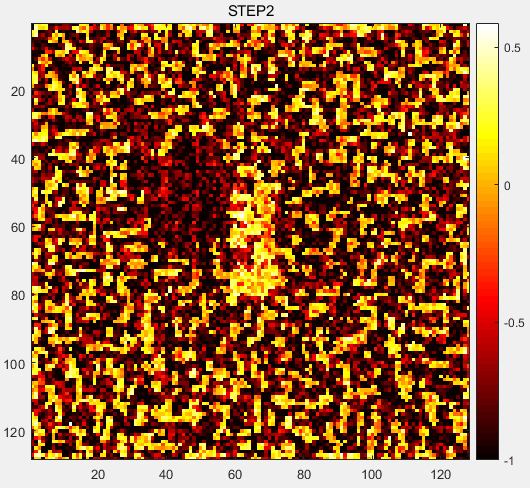}
\end{minipage}
\hspace{0.001\textwidth}
\begin{minipage}{0.31\textwidth}
\centering
\includegraphics[width=1.0\columnwidth]{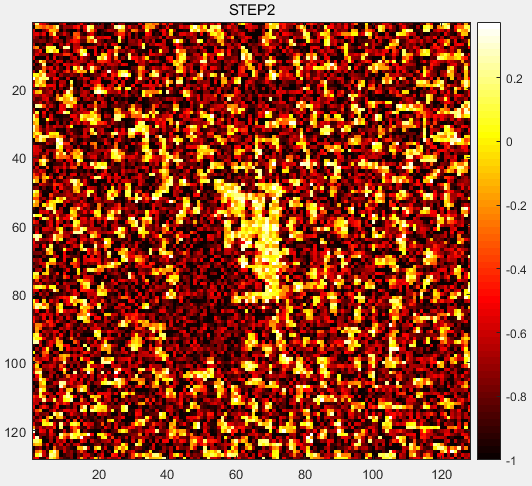}
\end{minipage}
}
\centerline{(b)}
\par{
\begin{minipage}{0.31\textwidth}
\centering
\includegraphics[width=1.0\columnwidth]{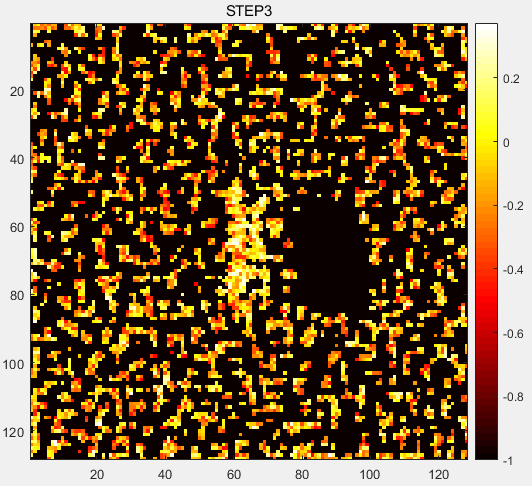}
\end{minipage}
\hspace{0.001\textwidth}
\begin{minipage}{0.31\textwidth}
\centering
\includegraphics[width=1.0\columnwidth]{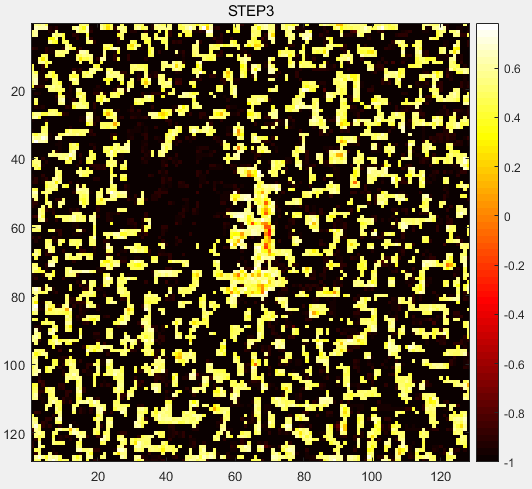}
\end{minipage}
\hspace{0.001\textwidth}
\begin{minipage}{0.31\textwidth}
\centering
\includegraphics[width=1.0\columnwidth]{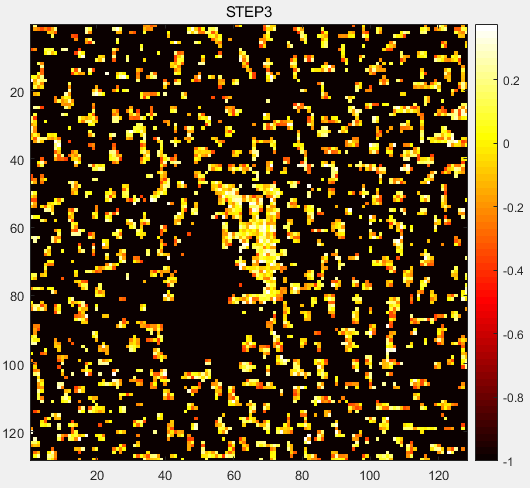}
\end{minipage}
}
\centerline{(c)}
\caption{Visualization of synaptic weights in the unsupervised learning bilayer SNN where each column from left to right is the feature map of BMP-2, BTR-60, T-72 via (a) 2, (b) 6, and (c) 20 training epochs, respectively.}
\label{fig:figure_placement}
\end{center}
\end{figure} 
In the unsupervised learning process of SNN, the synaptic weights are gradually sparse as the training continues. That is to say, each output layer neuron has a unique sensitivity to different receptive fields of the image. The smaller the synaptic weight is, the weaker the spike sequence transmits in the network. Compared to single layer SNN, the bilayer SNN has a faster convergence speed, and the value of synaptic weights is concentrated in a smaller range. It means that as SNN deepens, the classification performance is gradually enhanced. 
\begin{figure}[H]
\begin{center}
\par{
\begin{minipage}{0.32\textwidth}
\centering
\includegraphics[width=1.0\columnwidth]{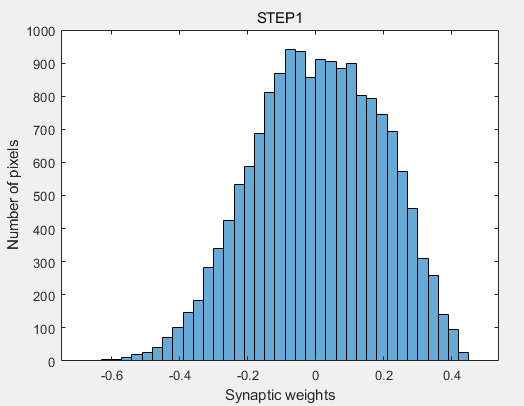}
\end{minipage}
\hspace{0.001\textwidth}
\begin{minipage}{0.32\textwidth}
\centering
\includegraphics[width=1.0\columnwidth]{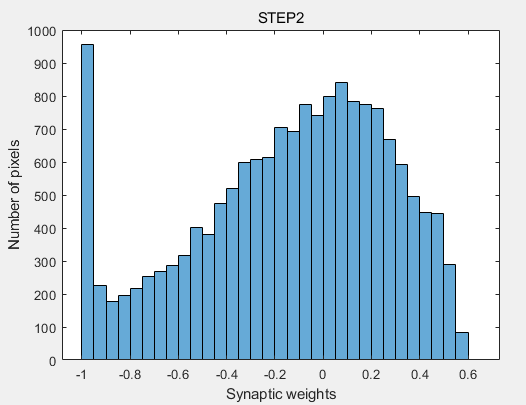}
\end{minipage}
\hspace{0.001\textwidth}
\begin{minipage}{0.32\textwidth}
\centering
\includegraphics[width=1.0\columnwidth]{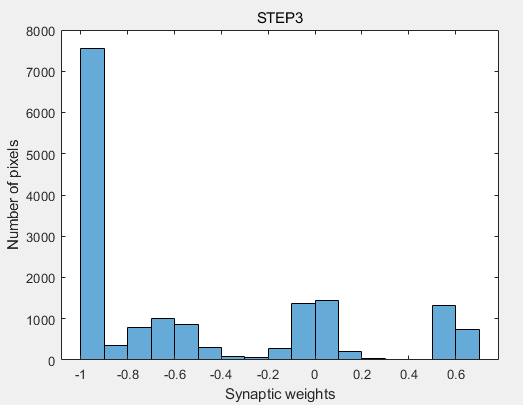}
\end{minipage}
}
\centerline{(a)}
\par{
\begin{minipage}{0.32\textwidth}
\centering
\includegraphics[width=1.0\columnwidth]{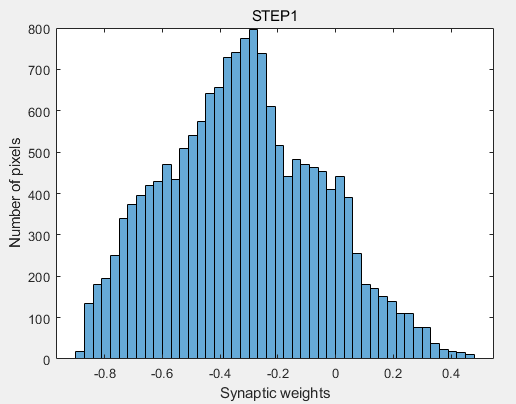}
\end{minipage}
\hspace{0.001\textwidth}
\begin{minipage}{0.32\textwidth}
\centering
\includegraphics[width=1.0\columnwidth]{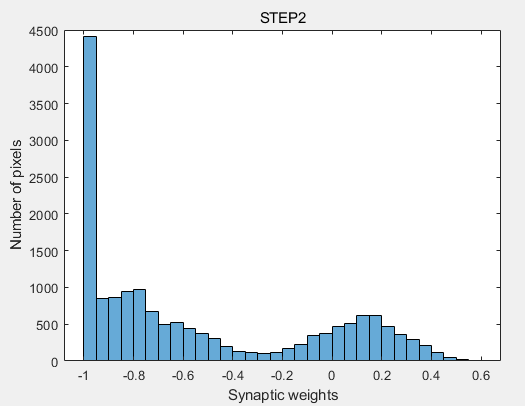}
\end{minipage}
\hspace{0.001\textwidth}
\begin{minipage}{0.32\textwidth}
\centering
\includegraphics[width=1.0\columnwidth]{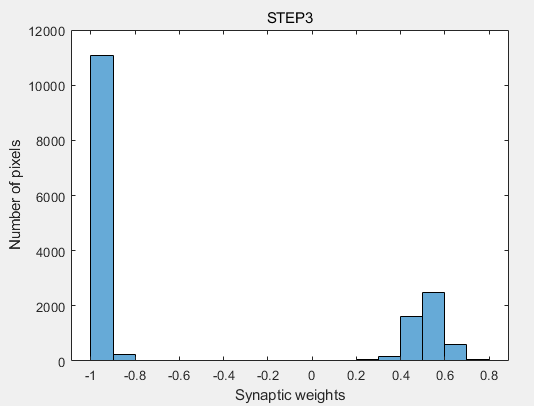}
\end{minipage}
}
\centerline{(b)}
\caption{Histograms of the synapse weight distribution in the feature map of BTR-60, where each column from left to right via 2, 6, and 20 training epochs and based on the unsupervised (a) single layer SNN and (b) bilayer SNN, respectively..}
\label{fig:figure_placement}
\end{center}
\end{figure} 

The histograms of the synapse weight distribution give a more intuitive description of the SNN’s sparseness, as shown in figure 13. The more synaptic weights are set to negative values (the sparser the SNN is), the stronger the feature extraction ability on the train set will be. However, it will cause overfitting if the training SNN is too sparse. Therefore, in actual tasks, the optimal model should be obtained through fine-tuning.

\subsection{SAR image classification based on supervised learning of SNN}
\subsubsection{Supervised learning algorithm of single layer SNN}
\begin{figure}[ht!]
\begin{center}
		\includegraphics[width=1.0\columnwidth]{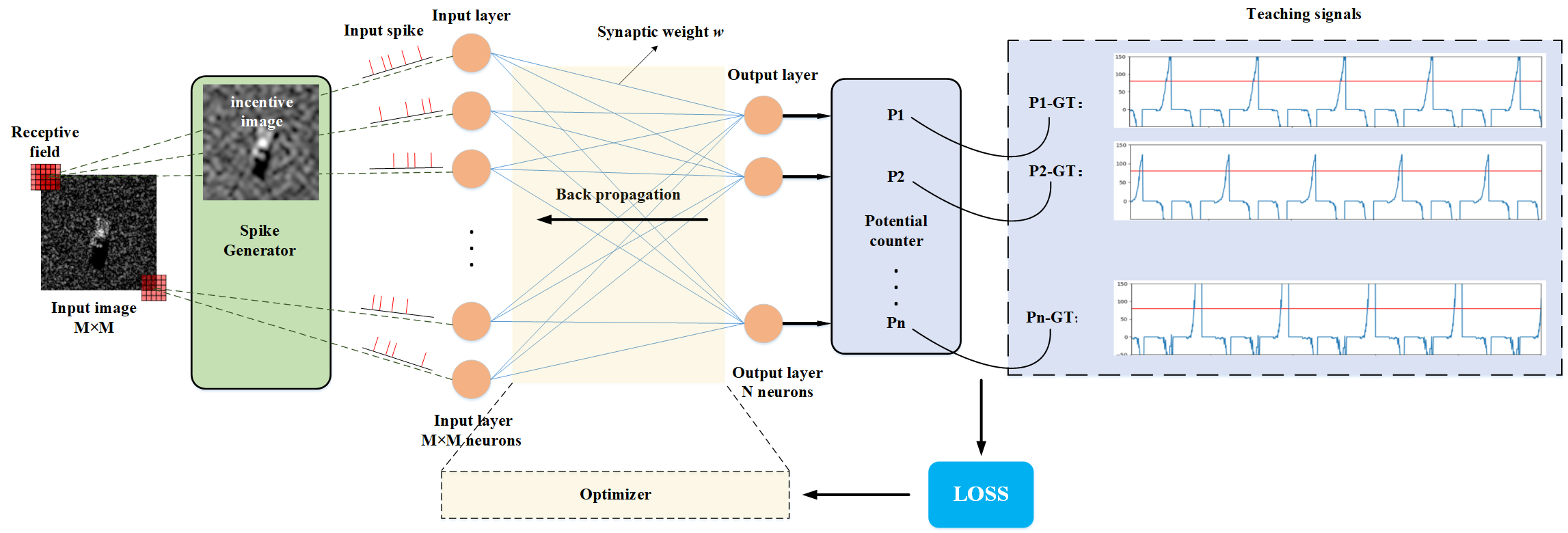}
	\caption{Architecture of supervised learning single layer SNN.}
\label{fig:figure_placement}
\end{center}
\end{figure}
Different supervised learning algorithms are suitable for various neural network structures. The main topological structures used include single neuron or single-layer network, multilayer feed-forward network, and recursive network. Generally, the more complex the system of SNN is, the more complicated the construction of a corresponding supervised learning algorithm becomes. We implement a bilayer supervised learning algorithm on SNN. The network structure is shown in figure 14. Our purpose is to minimize the difference between the target output membrane potential and the actual output membrane potential to modify synaptic weights. Firstly, the input image passes through the different receptive fields with Manhattan distance. Then the spike generator generates spike sequences and directly loaded them onto the first layer neurons. SNN transmits spike sequences through synapses. Finally, a loss function is designed to calculate the error between the actual output membrane potential and the guidance membrane potential. In the proposed SNN, we back propagate the error via optimizer.

To accelerate and optimize the algorithm implementation, we directly input spike sequences to the SNN. Those spike sequences (correspond to the input images) are obtained through SNN’s receptive field and spike generator. The guidance signal is the convergence membrane potential of output layer neurons based on the single layer unsupervised SNN (our previous work \cite{88}) . They are features that those neurons abstract from each category of the input image. Figure 15 shows the membrane potential guidance signals of BMP-2, BTR-60, and T-72, respectively. 
\begin{figure}[ht!]
\begin{center}
\par{
\begin{minipage}{0.9\textwidth}
\centering
\includegraphics[width=1.0\columnwidth]{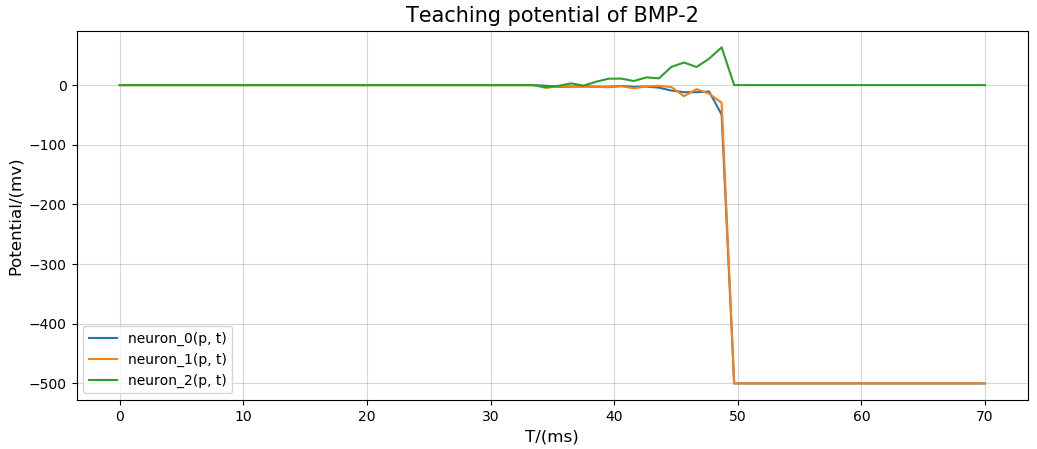}
\end{minipage}
}
\centerline{(a)}
\par{
\begin{minipage}{0.9\textwidth}
\centering
\includegraphics[width=1.0\columnwidth]{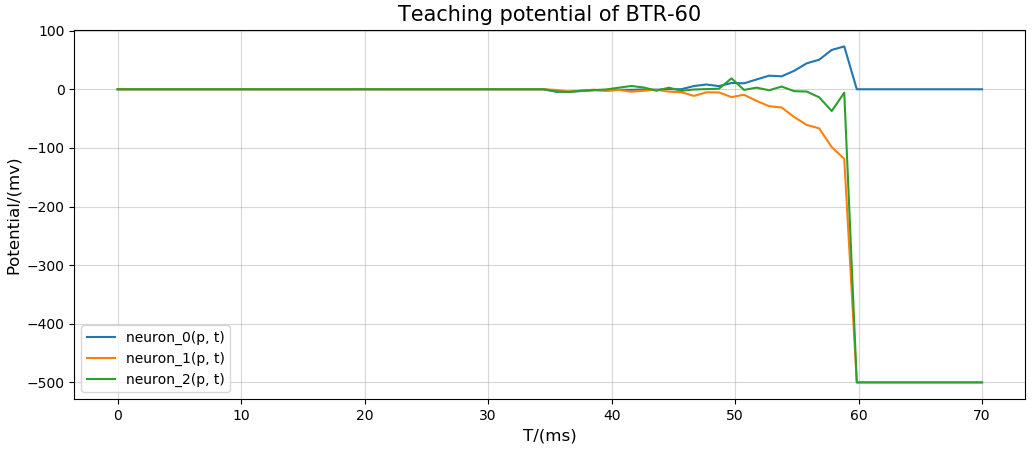}
\end{minipage}
}
\centerline{(b)}
\par{
\begin{minipage}{0.9\textwidth}
\centering
\includegraphics[width=1.0\columnwidth]{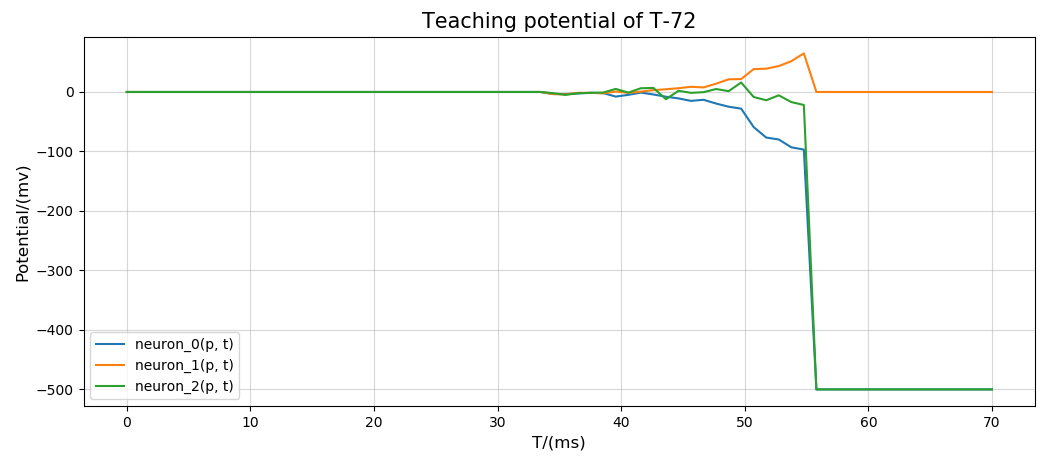}
\end{minipage}
}
\centerline{(c)}
\caption{Three kinds of membrane potential guidance signals (a) BMP-2, (b) BTR-60, and (c) T-72.}
\label{fig:figure_placement}
\end{center}
\end{figure}

We use the error between the actual output spike sequences and the guidance spike sequences to measure the accuracy of supervised learning. When the error is less than a given value, SNN’s learning iteration process will be ended. In the experiment, we apply Huber loss as the loss function. The Huber loss approach combines the advantages of the mean squared error (MSE) and the mean absolute error (MAE). As shown in formula (13), it is a piecewise-defined function:
 \begin{equation}
huber\_loss(y,\widehat y) = \left\{ {\begin{array}{*{20}{c}}
{\frac{1}{2}\sum\limits_{i = 1}^n {{{({y_i} - \widehat {{y_i}})}^2},{\rm{  }} \quad for\left| {{y_i} - \widehat {{y_i}}} \right| \le \delta } }\\
{\delta \sum\limits_{i = 1}^n {\left| {{y_i} - \widehat {{y_i}}} \right| - \frac{1}{2}{\delta ^2},{\rm{   }} \quad otherwise} }
\end{array}} \right\}
\end{equation}

Huber Loss is a robust regression loss function, where $\delta$ is a hyperparameter that controls the split between the two sub-function. The sub-function for large errors, such as outliers, is the absolute error function. Hence, it avoids the excessive sensitivity to large errors that characterize MSE. The sub-function for minor errors is the squared error making the whole function continuous and differentiable, which overcomes MAE’s convergence issues. In formula (13), $y_{i}$  is the observed value of the test sample, $\widehat y_{i}$ is the training sample statistics, and $\delta$ is set to 1 in the experiment.

Because of the multiplicative relationship between membrane potential, synaptic weight, and the input spike sequences, an internal chain rule is constructed by the optimizer to carry out error back-propagation. Our experiment uses the adaptive learning rate method (Adam) optimizer. The Adam algorithm calculates an exponential weighted moving average of the gradient and then squares the calculated gradient. It is computationally efficient and works well on problems with noisy or sparse gradients. Experimental results show that Adam significantly optimizes the SNN models better than stochastic gradient descent.

The performance of SNN is sensitive to its hyperparameters. Even for the bilayer SNN, the hyperparameters need to be debugged according to the LIF model‘s biological characteristics and the data set. The specific values of those hyperparameters are listed in table 3.
\begin{table}
\caption{Hyperparameters set for the supervised learning SNN.}\label{tab1}
\centering
\begin{tabular}{|l|l|l|}
\hline
Hyperparameters &  Symbol & Value\\
\hline
Number of neurons in the input layer &  $m$ & 16384\\
Number of neurons in the output layer &  $n$ & 3\\
Spike emission duration of a single pixel &  $T$ & 70 ms\\
Duration of refractory period & $t_{ref}$ & 20 ms\\
Resting potential &  $P_{rest}$ & 0.0 mv\\
Reset potential &  $P_{reset}$ & 0.0 mv\\
Threshold potential &  $P_{th}$ & 80.0 mv\\
Constant leakage potential &  $D$ & -5.0 mv\\
Transverse inhibitory potential &  $P_{in}$ & -500.0 mv\\
Number of batch samples &  $BATCH\_SIZE$ & 1\\
Maximum training steps &  $MAX\_STEPS$ & 30000\\
Initial learning rate&  $LR_{ini}$ & 1×10$^{-3}$\\
Mid-term learning rate &  $LR_{mid}$ & 1×10$^{-4}$\\
\hline
\end{tabular}
\end{table}

\subsubsection{Online processing ability and local learning characteristics}

In most cases, the data captured in the real world have temporal and spatial characteristics \cite{43}. The spatio-temporal data are generally represented as continuous spike sequence flow. Synaptic weights are required to modify dynamically along with the input spikes and real-time adapt to the data. Therefore, the online learning algorithm is more suitable and adequate for processing such real-time tasks \cite{44}.

The local characteristic means that the network’s learning rules are only determined by the spike sequence of the presynaptic and postsynaptic neurons. On the one hand, it indicates the scalability of the supervised learning algorithm when applied to a simple neural network or large-scale neural network. On the other hand, the supervised learning algorithm with local characteristics is suitable for implementing a parallel hardware system, which can solve specific problems efficiently.

\subsubsection{Classification results and feature visualization of synaptic weights}
In the experiment of SAR image classification based on supervised learning SNN, the overall classification accuracy versus training epoch is shown in figure 16. Based on the TensorFlow deep learning platform, we backpropagate error gradient by directly using optimizer and implements GPU acceleration. The overall classification accuracy is 70.43\% at the fifth training epoch and reaches 90.05\% after 25 training epochs. Our experimental platform is Ubuntu 18.04, 64G memory, and GPU Quadro RTX8000. When the input image size is 128 × 128, GPU’s occupation stables at 62\% and the model runs at 7 fps.

\begin{figure}[ht!]
\begin{center}
		\includegraphics[width=0.8\columnwidth]{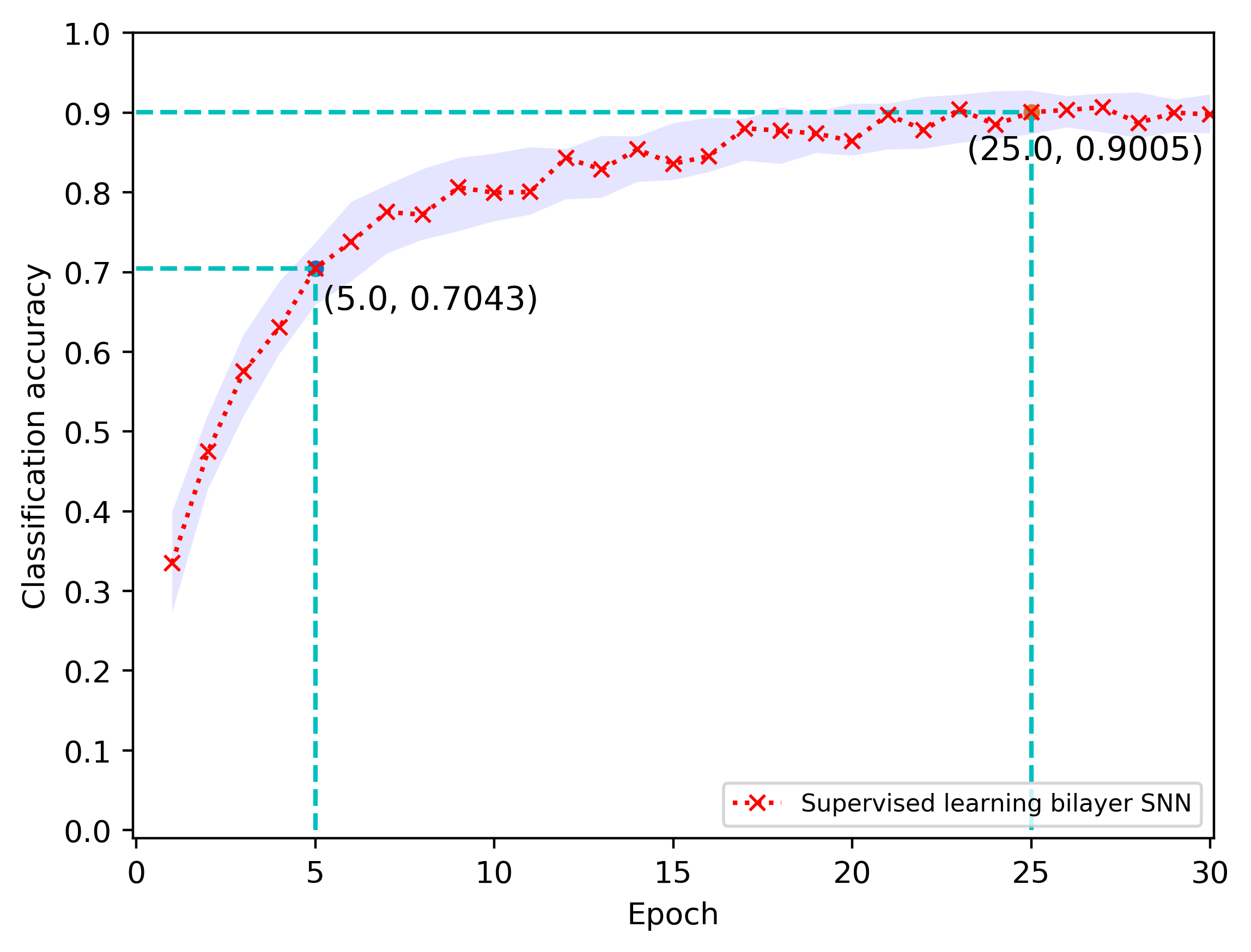}
	\caption{Classification accuracy versus training epoch of the supervised learning single layer SNN, where the purple shade represents the total performance range of the SNN under multiple repeated experiments.}
\label{fig:figure_placement}
\end{center}
\end{figure}

\begin{figure}[ht!]
\begin{center}
\par{
\begin{minipage}{0.31\textwidth}
\centering
\includegraphics[width=1.0\columnwidth]{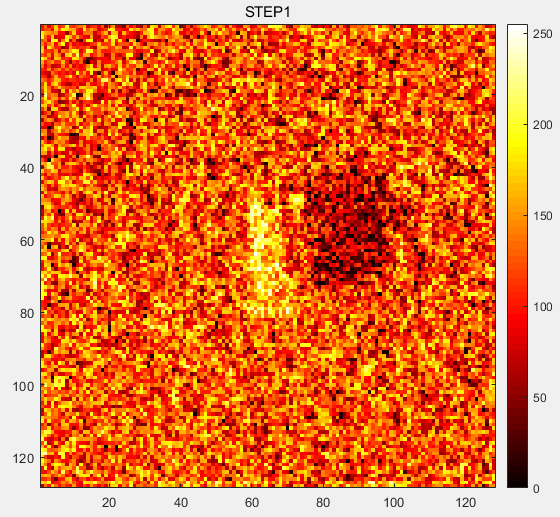}
\end{minipage}
\hspace{0.001\textwidth}
\begin{minipage}{0.31\textwidth}
\centering
\includegraphics[width=1.0\columnwidth]{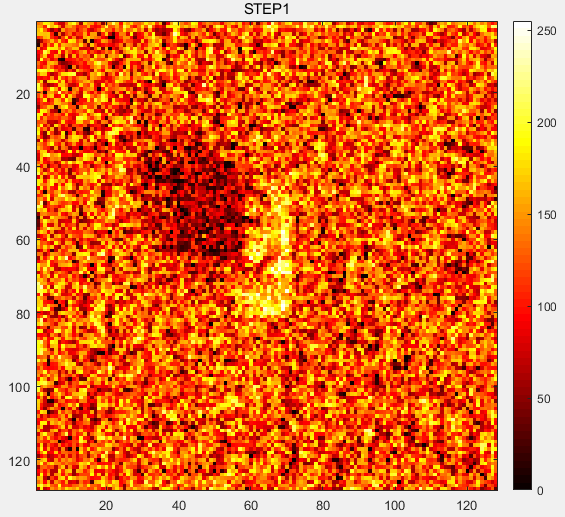}
\end{minipage}
\hspace{0.001\textwidth}
\begin{minipage}{0.31\textwidth}
\centering
\includegraphics[width=1.0\columnwidth]{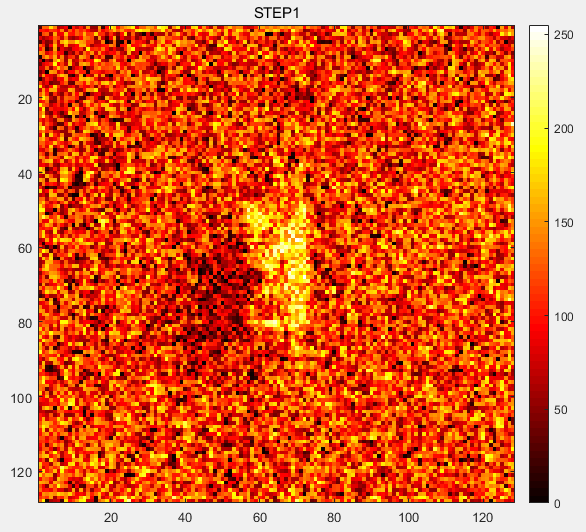}
\end{minipage}
}
\centerline{(a)}
\par{
\begin{minipage}{0.31\textwidth}
\centering
\includegraphics[width=1.0\columnwidth]{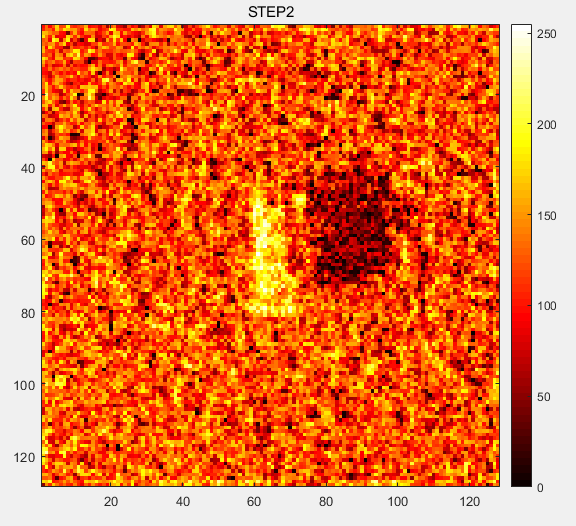}
\end{minipage}
\hspace{0.001\textwidth}
\begin{minipage}{0.31\textwidth}
\centering
\includegraphics[width=1.0\columnwidth]{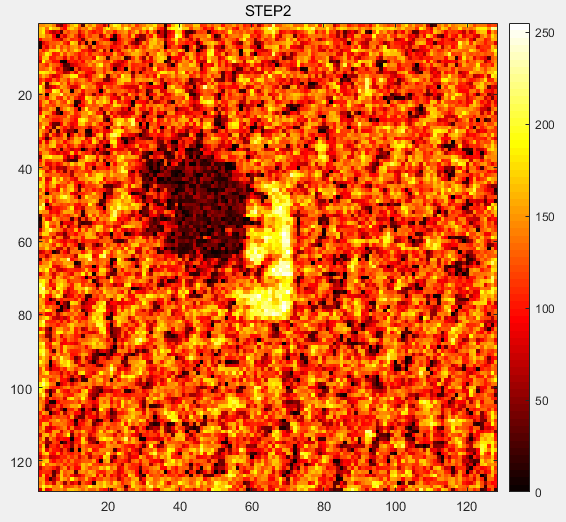}
\end{minipage}
\hspace{0.001\textwidth}
\begin{minipage}{0.31\textwidth}
\centering
\includegraphics[width=1.0\columnwidth]{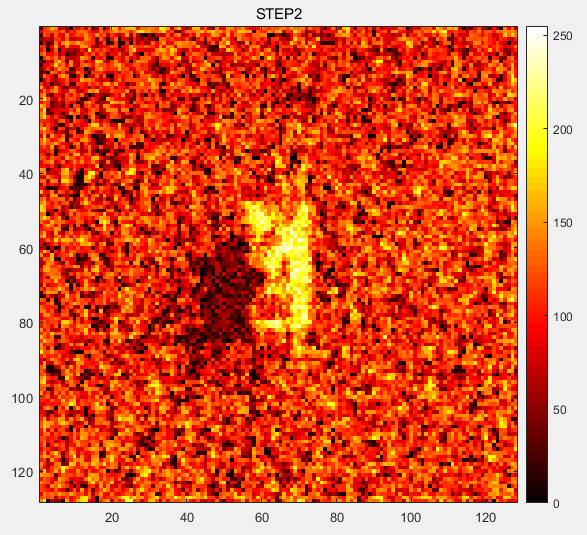}
\end{minipage}
}
\centerline{(b)}
\par{
\begin{minipage}{0.31\textwidth}
\centering
\includegraphics[width=1.0\columnwidth]{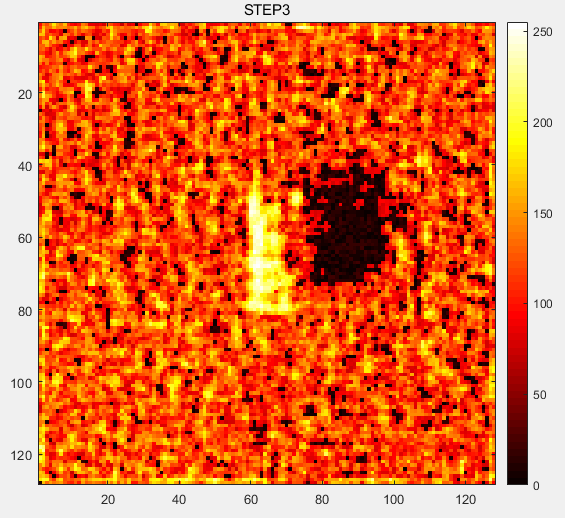}
\end{minipage}
\hspace{0.001\textwidth}
\begin{minipage}{0.31\textwidth}
\centering
\includegraphics[width=1.0\columnwidth]{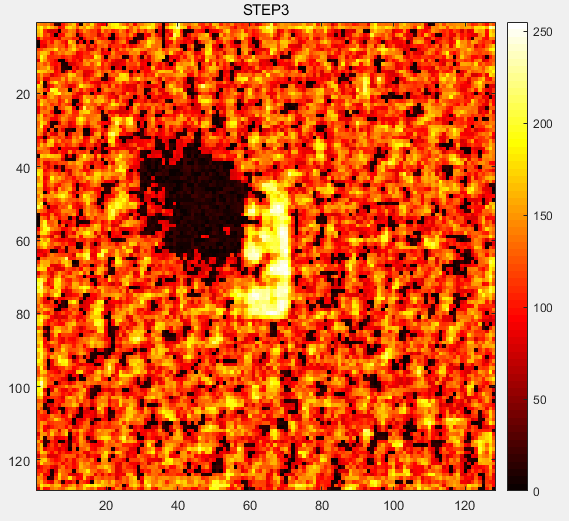}
\end{minipage}
\hspace{0.001\textwidth}
\begin{minipage}{0.31\textwidth}
\centering
\includegraphics[width=1.0\columnwidth]{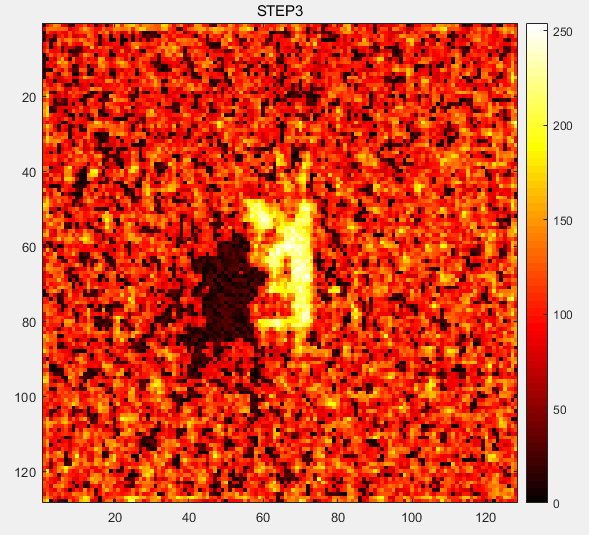}
\end{minipage}
}
\centerline{(c)}
\caption{Visualization of synaptic weights in the supervised learning bilayer SNN where each column from left to right is the feature map of BMP-2, BTR-60, T-72 via (a) 5, (b) 10, and (c) 25 training epochs, respectively.}
\label{fig:figure_placement}
\end{center}
\end{figure}
According to the receptive field, we visualize the SNN’s feature maps by rearranging synaptic weights (normalized to 0-255), as shown in figure 17. As can be seen from the visualization results, with the increasing training epoch, the synaptic weights constantly change towards different trends. That is, the sensitivities (or sensitive regions) of each output layer neuron versus the input image are different. The greater the synaptic weight is, the stronger the spike transmission will be. Moreover, the rapid variation of synaptic weights indicates that supervised SNN converges fast enough. With the increase of iterations, the feature maps can almost reconstruct the real image.

\begin{figure}[ht!]
\begin{center}
\begin{minipage}{0.60\textwidth}
\centering
\includegraphics[width=1.0\columnwidth]{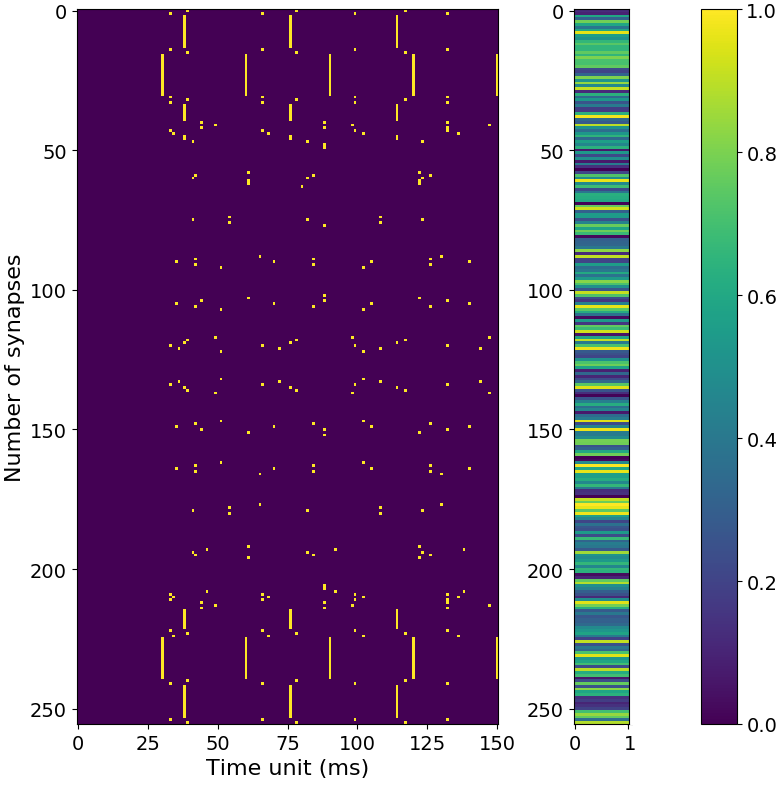}
\centerline{(a)}
\end{minipage}
\hspace{0.001\textwidth}
\begin{minipage}{0.80\textwidth}
\centering
\includegraphics[width=1.0\columnwidth]{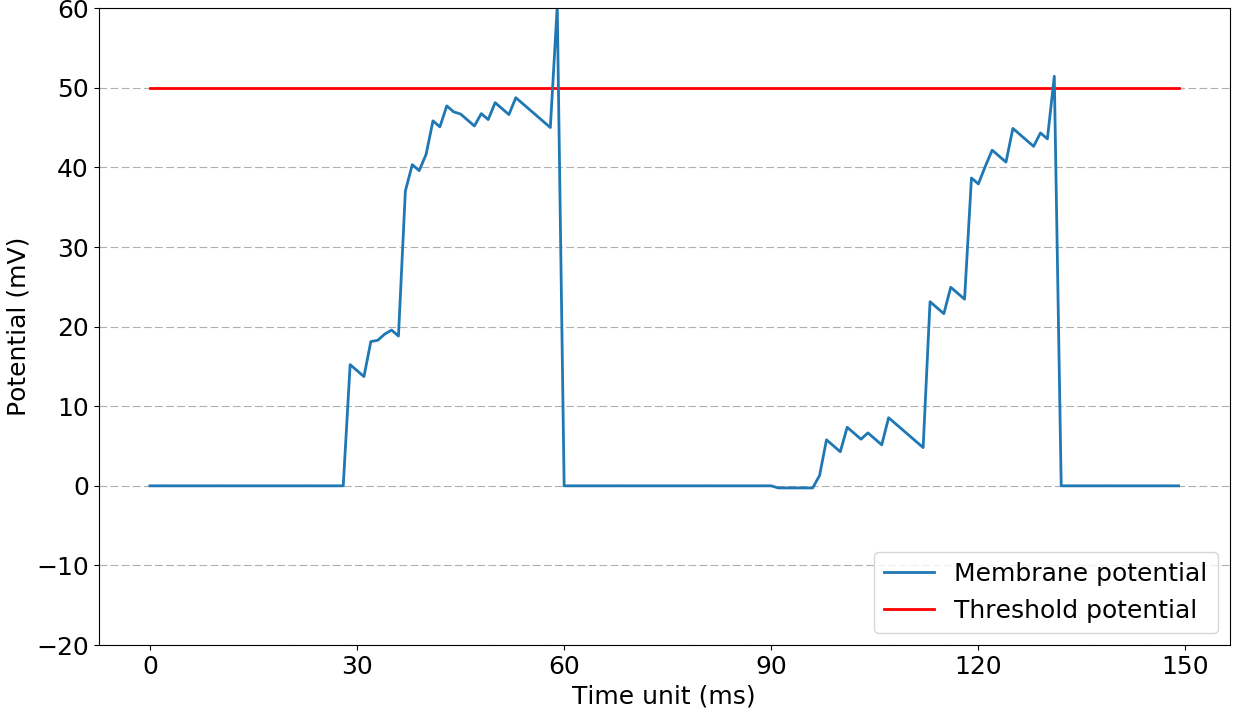}
\centerline{(b)}
\end{minipage}
\caption{Bionic simulation of the LIF neuron, (a) a set of input spike sequences and synaptic weights, (b) the corresponding postsynaptic neuron’s membrane potential versus time unit.}
\label{fig:figure_placement}
\end{center}
\end{figure}

\section{Analysis and evaluation of SNN}
\subsection{Bionic simulation of spiking neurons}
The SNN has rich neural computing characteristics, which is shown as the ability to simulate various spike emission activities of real biological neurons. We recorded a set of input spike sequences, synaptic weights, and the postsynaptic neuron’s membrane potential response guided by them, as shown in figure 18. We input an image with 16×16 to the SNN. In this process, the spike emission duration of a single pixel is 200 time units (200ms), and the neuron spike emission threshold is 50 mV.

Through the event-driven strategy, the bionic simulation of the LIF neuron is realized. We conclude that the spiking neuron model with a fixed threshold and reset mechanism has a higher level of abstraction. A linear differential equation can describe the dynamic response and has a unique stable equilibrium point in the resting state.

\subsection{Comparative analysis of SNN to CNN}
The power of CNN lies in the ability to learn features at multiple levels. The shallow convolution layer (with small receptive field) learns local features while the deeper convolution layer (with large receptive field) learns abstract features. Activation function decides, whether a neuron should be activated or not by calculating weighted sum and further adding bias with it. The activation function must be continuously differentiable, which causes CNN to be unable to solve complex problems such as real-time dynamic. 

So far, SNN is theoretically considered to be a more robust and biologically realistic model. The data processing of SNN adopts the spike sequence, including time, frequency, space, and other information, which significantly improves the computing ability and can process spatiotemporal information. Therefore, it can better simulate all kinds of neuron signals, be closer to the biological nervous system, and significantly improve biological authenticity. Inspired by related research \cite{45}, we explore the advantages of SNN's noise resistance in image classification.

We add white noise to the dataset artificially according to different SNRs. The preprocessed samples are shown in figure 19. We compare the performance between SNN and CNN when images are in the presence of noise. 

\begin{figure}[ht!]
\begin{center}
\par{
\begin{minipage}{0.185\textwidth}
\centering
\includegraphics[width=1.0\columnwidth]{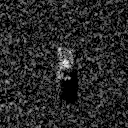}
\end{minipage}
\hspace{0.0001\textwidth}
\begin{minipage}{0.185\textwidth}
\centering
\includegraphics[width=1.0\columnwidth]{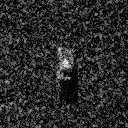}
\end{minipage}
\hspace{0.0001\textwidth}
\begin{minipage}{0.185\textwidth}
\centering
\includegraphics[width=1.0\columnwidth]{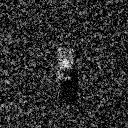}
\end{minipage}
\hspace{0.0001\textwidth}
\begin{minipage}{0.185\textwidth}
\centering
\includegraphics[width=1.0\columnwidth]{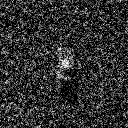}
\end{minipage}
\hspace{0.0001\textwidth}
\begin{minipage}{0.185\textwidth}
\centering
\includegraphics[width=1.0\columnwidth]{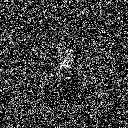}
\end{minipage}
}
\centerline{(a)}
\par{
\begin{minipage}{0.185\textwidth}
\centering
\includegraphics[width=1.0\columnwidth]{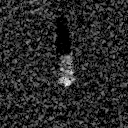}
\end{minipage}
\hspace{0.0001\textwidth}
\begin{minipage}{0.185\textwidth}
\centering
\includegraphics[width=1.0\columnwidth]{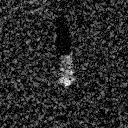}
\end{minipage}
\hspace{0.0001\textwidth}
\begin{minipage}{0.185\textwidth}
\centering
\includegraphics[width=1.0\columnwidth]{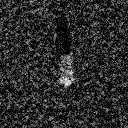}
\end{minipage}
\hspace{0.0001\textwidth}
\begin{minipage}{0.185\textwidth}
\centering
\includegraphics[width=1.0\columnwidth]{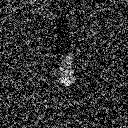}
\end{minipage}
\hspace{0.0001\textwidth}
\begin{minipage}{0.185\textwidth}
\centering
\includegraphics[width=1.0\columnwidth]{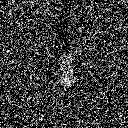}
\end{minipage}
}
\centerline{(b)}
\par{
\begin{minipage}{0.185\textwidth}
\centering
\includegraphics[width=1.0\columnwidth]{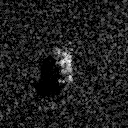}
\end{minipage}
\hspace{0.0001\textwidth}
\begin{minipage}{0.185\textwidth}
\centering
\includegraphics[width=1.0\columnwidth]{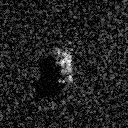}
\end{minipage}
\hspace{0.0001\textwidth}
\begin{minipage}{0.185\textwidth}
\centering
\includegraphics[width=1.0\columnwidth]{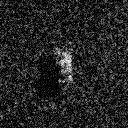}
\end{minipage}
\hspace{0.0001\textwidth}
\begin{minipage}{0.185\textwidth}
\centering
\includegraphics[width=1.0\columnwidth]{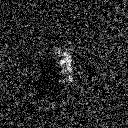}
\end{minipage}
\hspace{0.0001\textwidth}
\begin{minipage}{0.185\textwidth}
\centering
\includegraphics[width=1.0\columnwidth]{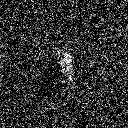}
\end{minipage}
}
\centerline{(c)}
\caption{Samples of adding noise according to different SNRs, (a) BMP-2, (b) BTR-60 and (c) T-72. Each column from left to right are original images, images with SNR of 10dB, images with SNR of 5dB, images with SNR of 0dB, and images with SNR of -5dB.}
\label{fig:figure_placement}
\end{center}
\end{figure}
In the noise resistance experiment, we train the proposed single layer supervised SNN and a comparable shallow CNN with a convolution layer, a pooling layer, and a fully connected layer. Both of them are supervised and using pre-trained model (traning samples without artificial noise). The overall classification accuracy of the networks versus SNR is shown in figure 20. The SNN is superior to the CNN in noise resistance ability, and can be effective in SAR image classification with severe speckle noise.

\begin{figure}[ht!]
\begin{center}
		\includegraphics[width=0.9\columnwidth]{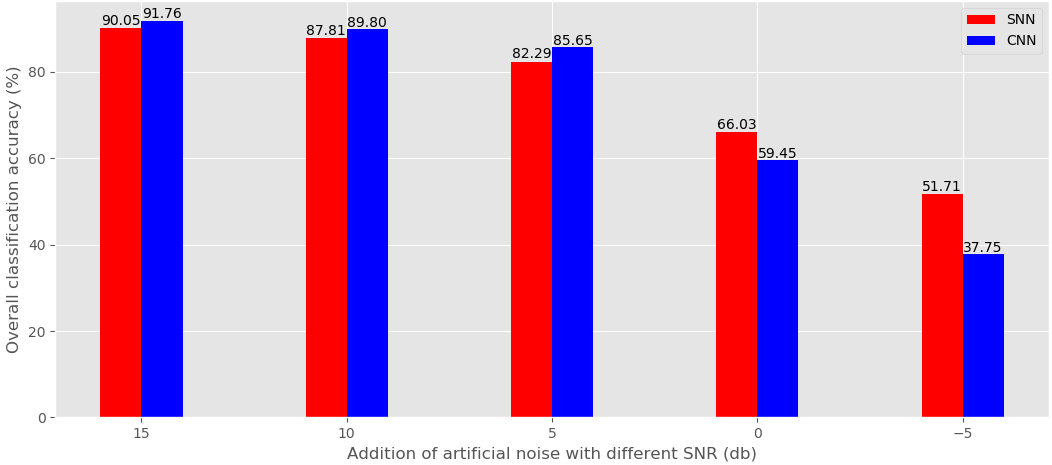}
	\caption{The overall classification accuracy of the shallow supervised SNN and CNN versus addition of artificial noise with different SNR.}
\label{fig:figure_placement}
\end{center}
\end{figure}

In addition to the classification accuracy, additional evaluation indexes are needed to evaluate the model: 1) GFLOPs (gigabyte floating-point operations), which can be understood as a calculation amount required for forward propagation and reflects the hardware's performance requirements such as GPU. 2) Number of parameters describes model complexity. 3) Model size reflects the amount of memory occupied. 4) GPU occupancy in training characterizes the computational efficiency of the training process. 5) Model speed is used to measure the time required for a model to train or test a single image. We list the experiment measurement results in table 4. The above indicators have an essential reference for model implementation.
\begin{table}[H]
	\centering
		\begin{tabular}{|l|c|c|}\hline
			Evaluation index /&CNN&SNN\\
			Network structure&(a convolution layer,&(single-layer,\\
			&a pooling layer,&supervised learning)\\
			&a fully connected layer,&\\
			&supervised learning)&\\\hline
			 Gigabyte floating-&5.15&9.83×10$^{-5}$\\
             point operations (GFLOPs)&&\\\hline			 
			 Number of parameters (M)&1.58&4.90×10$^{-2}$\\\hline
			 Model size (MB)&6.183&1.967×10$^{-1}$\\\hline
			 GPU occupancy&95&62\\
			 in training (\%)&&\\\hline
			 Model speed (FPS)&42.75&7.21\\\hline
		\end{tabular}
	\caption{Model evaluation of SNN and CNN.}
\label{tab:Partial MSTAR dataset}
\end{table}

Based on the TensorFlow deep learning platform, we backpropagate the error gradient of SNN by directly using optimizer and implements GPU acceleration. Our experimental platform is Ubuntu 18.04, 64G memory, and GPU Quadro RTX8000. When the input image size is 128×128, GPU’s occupation stables at 62$\%$, and with the fps $\approx$ 7.

The FLOPs of SNN is much less than that of CNN. So the SNN consumes less computing resources, which means that it is more conducive to layout algorithms on small-scale hardware architectures such as vehicle or airborne. Compared with CNN, SNN also has obvious advantages in the number of parameters and the model size. During the training process, the SNN’s GPU occupancy also remained low, proving that SNN has the characteristics of low energy consumption in engineering applications. In our current work, SNN runs not as fast as CNN. This is because a large number of tensor judgment operations in the underlying SNN’s construction are considered comprehensively. We have yet to develop more optimization algorithms. As far as we know, the newly developing gradient substitution method and the ANN2SNN method on the SpikingJelly \cite{46} platform provide the possibility for the acceleration of SNN. In summary, the computational efficiency of the SNN (transmitting the time-domain information of spike sequences) is better than that of the CNN (using Sigmoid as activation function). Moreover, SNN calculates discrete spike sequence instead of analog signal, which is more suitable for hardware implementation and processing.

\section{Conclusion}

This article proposes a full-link method from the unsupervised learning of SNN based on STDP to the supervised learning of SNN based on gradient descent. The contribution of the method is to expose spike sequence processing in the whole neural network. One of the most innovative is the single-layer supervised SNN, which has been realized based on the TensorFlow platform. The classification accuracy reaches 90.05\% in three categories of images on the MSTAR dataset, making the SNN basically comparable with CNN. The visualization of synapse weights also certifies the reliability and accuracy of SNN in image classification tasks. 



\begin{thebibliography}{10}
	\expandafter\ifx\csname url\endcsname\relax
	\def\url#1{\texttt{#1}}\fi
	\expandafter\ifx\csname urlprefix\endcsname\relax\def\urlprefix{URL }\fi
	\expandafter\ifx\csname href\endcsname\relax
	\def\href#1#2{#2} \def\path#1{#1}\fi
	
	\bibitem{1}
	W.~Mcculloch, W.~Pitts, A logical calculus of the ideas immanent in nervous
	activity, The bulletin of mathematical biophysics 5~(4) (1943) 115--133.
	\newblock \href {http://dx.doi.org/10.1007/BF02478259}
	{\path{doi:10.1007/BF02478259}}.
	
	\bibitem{2}
	D.~E. Rumelhart, G.~E. Hinton, R.~J. Williams, Learning representations by
	back-propagating errors, Nature 323~(6088) (1988) 533--536.
	\newblock \href {http://dx.doi.org/10.1038/323533a0}
	{\path{doi:10.1038/323533a0}}.
	
	\bibitem{3}
	A.~K. Engel, P.~Konig, et~al., Temporal coding in the visual cortex: new vistas
	on integration in the nervous system - sciencedirect, Trends in Neurosciences
	15~(6) (1992) 218--226.
	\newblock \href {http://dx.doi.org/10.1016/0166-2236(92)90039-B}
	{\path{doi:10.1016/0166-2236(92)90039-B}}.
	
	\bibitem{4}
	K.~I. Mcanally, J.~F. Stein, Auditory temporal coding in dyslexia, Proceedings
	of the Royal Society B: Biological Sciences 263~(1373) (1996) 961--965.
	\newblock \href {http://dx.doi.org/10.1098/rspb.1996.0142}
	{\path{doi:10.1098/rspb.1996.0142}}.
	
	\bibitem{5}
	A.~K. Seth, Neural coding: Rate and time codes work together, Current Biology
	25~(3) (2015) 110--113.
	\newblock \href {http://dx.doi.org/10.1016/j.cub.2014.12.043}
	{\path{doi:10.1016/j.cub.2014.12.043}}.
	
	\bibitem{6}
	E.~M. Izhikevich, Which model to use for cortical spiking neurons?, IEEE
	Transactions on Neural Networks 15~(5) (2004) 1063--1070.
	\newblock \href {http://dx.doi.org/10.1109/TNN.2004.832719}
	{\path{doi:10.1109/TNN.2004.832719}}.
	
	\bibitem{7}
	F.~Theunissen, J.~Miller, Temporal encoding in nervous systems: A rigorous
	definition, Journal of Computational Neuroscience 2~(2) (1995) 149--162.
	\newblock \href {http://dx.doi.org/10.1007/BF00961885}
	{\path{doi:10.1007/BF00961885}}.
	
	\bibitem{8}
	N.~Kasabov, N.~M. Scott, E.~Tu, et~al., Evolving spatio-temporal data machines
	based on the neucube neuromorphic framework: Design methodology and selected
	applications, Neural Networks 2 (2016) 1--14.
	\newblock \href {http://dx.doi.org/10.1016/j.neunet.2015.09.011}
	{\path{doi:10.1016/j.neunet.2015.09.011}}.
	
	\bibitem{9}
	S.~M. Bohte, H.~L. Poutre, J.~N. Kok, Unsupervised clustering with spiking
	neurons by sparse temporal coding and multilayer {RBF} networks, IEEE
	Transactions on Neural Networks 13~(2) (2002) 426--435.
	\newblock \href {http://dx.doi.org/10.1109/72.991428}
	{\path{doi:10.1109/72.991428}}.
	
	\bibitem{10}
	P.~U. Diehl, M.~Cook, Unsupervised learning of digit recognition using
	spike-timing-dependent plasticity, Frontiers in Computational Neuroscience
	9~(429) (2015) 99.
	\newblock \href {http://dx.doi.org/10.3389/fncom.2015.00099}
	{\path{doi:10.3389/fncom.2015.00099}}.
	
	\bibitem{11}
	S.~R. Kheradpisheh, M.~Ganjtabesh, T.~Masquelier, Bio-inspired unsupervised
	learning of visual features leads to robust invariant object recognition,
	Neurocomputing 205 (2016) 382--392.
	\newblock \href {http://dx.doi.org/10.1016/j.neucom.2016.04.029}
	{\path{doi:10.1016/j.neucom.2016.04.029}}.
	
	\bibitem{12}
	E.~I. Knudsen, Supervised learning in the brain., Journal of Neuroscience
	14~(7) (1994) 3985--3997.
	\newblock \href {http://dx.doi.org/10.1097/00005072-199407000-00013}
	{\path{doi:10.1097/00005072-199407000-00013}}.
	
	\bibitem{13}
	S.~M. Bohte, J.~N. Kok, H.~L. Poutré, Error-backpropagation in temporally
	encoded networks of spiking neurons, Neurocomputing 48~(1–4) (2002) 17--37.
	\newblock \href {http://dx.doi.org/10.1016/S0925-2312(01)00658-0}
	{\path{doi:10.1016/S0925-2312(01)00658-0}}.
	
	\bibitem{14}
	W.~Gerstner, W.~M. Kistler, Spiking neuron models: {Single Neurons},
	{Populations}, {Plasticity}, Cambridge University Press. \,\href
	{http://dx.doi.org/10.1017/CBO9780511815706}
	{\path{doi:10.1017/CBO9780511815706}}.
	
	\bibitem{15}
	S.~Mckennoch, D.~Liu, L.~G. Bushnell, Fast modifications of the spikeprop
	algorithm, in: Proceedings of the International Joint Conference on Neural
	Networks, IEEE, 2006, pp. 3970--3977.
	\newblock \href {http://dx.doi.org/10.1109/IJCNN.2006.246918}
	{\path{doi:10.1109/IJCNN.2006.246918}}.
	
	\bibitem{16}
	S.~Mckennoch, T.~Voegtlin, L.~Bushnell, Spike-timing error backpropagation in
	theta neuron networks., Neural Computation 21~(1) (2009) 9--45.
	\newblock \href {http://dx.doi.org/10.1162/neco.2009.09-07-610}
	{\path{doi:10.1162/neco.2009.09-07-610}}.
	
	\bibitem{17}
	S.~Ghosh-Dastidar, H.~Adeli, A new supervised learning algorithm for multiple
	spiking neural networks with application in epilepsy and seizure detection,
	Neural networks: the official journal of the International Neural Network
	Society 22~(10) (2009) 1419--1431.
	\newblock \href {http://dx.doi.org/10.1016/j.neunet.2009.04.003}
	{\path{doi:10.1016/j.neunet.2009.04.003}}.
	
	\bibitem{18}
	Y.~Xu, X.~Zeng, L.~Han, J.~Yang, A supervised multi-spike learning algorithm
	based on gradient descent for spiking neural networks, Neural Networks 43~(4)
	(2013) 99--113.
	\newblock \href {http://dx.doi.org/10.1016/j.neunet.2013.02.003}
	{\path{doi:10.1016/j.neunet.2013.02.003}}.
	
	\bibitem{19}
	W.~Gerstner, Time structure of the activity in neural network models, Phys Rev
	E Stat Phys Plasmas Fluids Relat Interdiscip Topics 51~(1) (1995) 738--758.
	\newblock \href {http://dx.doi.org/10.1103/PhysRevE.51.738}
	{\path{doi:10.1103/PhysRevE.51.738}}.
	
	\bibitem{20}
	W.~M. Kistler, W.~Gerstner, J.~Hemmen, Reduction of the hodgkin-huxley
	equations to a single-variable threshold model, Neural Computation 9~(5)
	(1997) 1015--1045.
	\newblock \href {http://dx.doi.org/10.1162/neco.1997.9.5.1015}
	{\path{doi:10.1162/neco.1997.9.5.1015}}.
	
	\bibitem{21}
	P.~Tino, A.~J.~S. Mills, Learning beyond finite memory in recurrent networks of
	spiking neurons, Neural Computation 18~(3) (2006) 591--613.
	\newblock \href {http://dx.doi.org/10.1162/neco.2006.18.3.591}
	{\path{doi:10.1162/neco.2006.18.3.591}}.
	
	\bibitem{22}
	F.~Xu, H.~P. Wang, Y.~Q. Jin, Deep learning as applied in sar target
	recognition and terrain classification, Journal of Radars 6~(2) (2017)
	136--148.
	\newblock \href {http://dx.doi.org/info:doi/10.12000/JR16130}
	{\path{doi:info:doi/10.12000/JR16130}}.
	
	\bibitem{23}
	X.~X. Zhu, D.~Tuia, et~al., Deep learning in remote sensing: A comprehensive
	review and list of resources, IEEE Geoscience and Remote Sensing Magazine
	5~(4) (2017) 8--36.
	\newblock \href {http://dx.doi.org/10.1109/MGRS.2017.2762307}
	{\path{doi:10.1109/MGRS.2017.2762307}}.
	
	\bibitem{24}
	H.~Xie, W.~Shuang, et~al., Multilayer feature learning for polarimetric
	synthetic radar data classification, in: IEEE International Geoscience and
	Remote Sensing Symposium, 2014, pp. 2818--2821.
	\newblock \href {http://dx.doi.org/10.1109/IGARSS.2014.6947062}
	{\path{doi:10.1109/IGARSS.2014.6947062}}.
	
	\bibitem{25}
	Q.~Lv, Y.~Dou, et~al., Urban land use and land cover classification using
	remotely sensed sar data through deep belief networks, Journal of Sensors
	2015 (2015) 1--10.
	\newblock \href {http://dx.doi.org/10.1155/2015/538063}
	{\path{doi:10.1155/2015/538063}}.
	
	\bibitem{26}
	C.~Bentes, D.~Velotto, B.~Tings, Ship classification in terrasar-x images with
	convolutional neural networks, IEEE Journal of Oceanic Engineering PP~(99)
	(2017) 1--9.
	\newblock \href {http://dx.doi.org/10.1109/JOE.2017.2767106}
	{\path{doi:10.1109/JOE.2017.2767106}}.
	
	\bibitem{27}
	Z.~Peng, L.~Ming, et~al., Unsupervised multi-class segmentation of sar images
	using fuzzy triplet markov fields model, Pattern Recognition 32~(11) (2011)
	1532--1540.
	\newblock \href {http://dx.doi.org/10.1016/j.patrec.2011.04.009}
	{\path{doi:10.1016/j.patrec.2011.04.009}}.
	
	\bibitem{28}
	S.~Gupta, A.~Vyas, Spiking-neural-network, GitHub repository, Dec. 31, 2018,
	[Online] Available:
	\url{https://github.com/Shikhargupta/Spiking-Neural-Network}.
	
	\bibitem{29}
	L.~Lapicque, Recherches quantitatives sur l’excitation electrique des nerfs
	trainee comme une polarisation, Journal of Physiology and Pathololgy 9 (1907)
	620--635.
	
	\bibitem{32}
	D.~O. Hebb, The Organization Of Behavior A Neuropsychological Theory, New York:
	Wiley Press, 1949.
	
	\bibitem{33}
	H.~Markram, J.~Lübke, et~al., Regulation of synaptic efficacy by coincidence
	of postsynaptic aps and epsps, Science 275~(5297) (1997) 213--215.
	\newblock \href {http://dx.doi.org/10.1126/science.275.5297.213}
	{\path{doi:10.1126/science.275.5297.213}}.
	
	\bibitem{34}
	J.~Sjöström, W.~Gerstner, Spike-timing dependent plasticity, Scholarpedia
	5~(2) (2010) 1362.
	\newblock \href {http://dx.doi.org/10.4249/scholarpedia.1362}
	{\path{doi:10.4249/scholarpedia.1362}}.
	
	\bibitem{36}
	W.~M. Kistler, J.~L. Hemmen, Modeling synaptic plasticity in conjunction with
	the timing of pre- and postsynaptic action potentials, Neural Computation
	12~(2) (2000) 385--405.
	\newblock \href {http://dx.doi.org/10.1162/089976600300015844}
	{\path{doi:10.1162/089976600300015844}}.
	
	\bibitem{37}
	Z.~Wang, N.~L. Xu, et~al., Bidirectional changes in spatial dendritic
	integration accompanying long-term synaptic modifications., Neuron 37~(3)
	(2003) 463--472.
	\newblock \href {http://dx.doi.org/10.1016/S0896-6273(02)01189-3}
	{\path{doi:10.1016/S0896-6273(02)01189-3}}.
	
	\bibitem{41}
	E.~R. Keydel, S.~W. Lee, J.~T. Moore, {MSTAR} extended operating conditions:
	{A} tutorial, Proceedings of SPIE - The International Society for Optical
	Engineering (1996) 228–242\href {http://dx.doi.org/10.1117/12.242059}
	{\path{doi:10.1117/12.242059}}.
	
	\bibitem{88}
	J.~Chen, X.~Qiu, et~al., Unsupervised learning method for sar image
	classification based on spiking neural network, EUSAR Preprints 2021.\,\href
	{http://dx.doi.org/10.20944/preprints202102.0083.v1}
	{\path{doi:10.20944/preprints202102.0083.v1}}.
	
	\bibitem{43}
	N.~K. Kasabov, Neucube: A spiking neural network architecture for mapping,
	learning and understanding of spatio-temporal brain data, Neural Networks
	52~(4) (2014) 62--76.
	\newblock \href {http://dx.doi.org/10.1016/j.neunet.2014.01.006}
	{\path{doi:10.1016/j.neunet.2014.01.006}}.
	
	\bibitem{44}
	Y.~Oniz, O.~Kaynak, Variable-structure-systems based approach for online
	learning of spiking neural networks and its experimental evaluation, Journal
	of the Franklin Institute 351~(6) (2014) 3269--3285.
	\newblock \href {http://dx.doi.org/10.1016/j.jfranklin.2014.03.002}
	{\path{doi:10.1016/j.jfranklin.2014.03.002}}.
	
	\bibitem{45}
	T.~L. Zhang, Y.~Zeng, et~al., Hmsnn: Hippocampus inspired memory spiking neural
	network, in: IEEE International Conference on Systems, Man, and Cybernetics
	(SMC), 2016, p. 2301–2306.
	\newblock \href {http://dx.doi.org/10.1109/SMC.2016.7844581}
	{\path{doi:10.1109/SMC.2016.7844581}}.
	
	\bibitem{46}
	F.~Wang, et~al., Spikingjelly, GitHub repository, Mar. 12, 2021, [Online]
	Available: \url{https://github.com/fangwei123456/ spikingjelly}.
	
\end{thebibliography}
\end{document}